\documentclass[sigconf]{acmart}

\AtBeginDocument{%
  \providecommand\BibTeX{{%
    \normalfont B\kern-0.5em{\scshape i\kern-0.25em b}\kern-0.8em\TeX}}}

\copyrightyear{2021} 
\acmYear{2021} 
\setcopyright{rightsretained} 
\acmConference[AISec '21]{Proceedings of the 14th ACM Workshop on Artificial Intelligence and Security}{November 15, 2021}{Virtual Event, Republic of Korea}
\acmBooktitle{Proceedings of the 14th ACM Workshop on Artificial Intelligence and Security (AISec '21), November 15, 2021, Virtual Event, Republic of Korea}\acmDOI{10.1145/3474369.3486878}
\acmISBN{978-1-4503-8657-9/21/11}

\usepackage[T1]{fontenc}

\usepackage{graphicx}
\usepackage{comment}
\usepackage{amsmath} %
\usepackage{color}
\usepackage{bm}
\usepackage{xspace}
\usepackage{mathtools}
\usepackage{enumitem}
\usepackage{booktabs}
\usepackage{multirow}
\usepackage[bottom]{footmisc}
\usepackage{subcaption}
\usepackage{wrapfig}
\usepackage{soul}
\usepackage{amsthm}
\usepackage{nicefrac}       %
\usepackage{microtype}
\usepackage{amsbsy}
\usepackage{bbm}
\usepackage{stfloats}
\usepackage[linesnumbered,ruled,shortend]{algorithm2e}
\usepackage{mathrsfs}
\usepackage{thmtools}
\usepackage{thm-restate}
\usepackage{cleveref}
\usepackage{xr}

\makeatletter
\newcommand{\removelatexerror}{\let\@latex@error\@gobble}
\makeatother

\makeatletter
\def\th@plain{%
  \thm@notefont{}%
  \itshape %
}
\def\th@definition{%
  \thm@notefont{}%
  \normalfont %
}
\makeatother

\newcommand{\norm}[1]{\left\lVert#1\right\rVert}
\DeclareMathOperator*{\argmin}{arg\,min}
\DeclareMathOperator*{\argmax}{arg\,max}

\DeclarePairedDelimiter{\iprod}{\langle}{\rangle}
\DeclarePairedDelimiter\abs{\lvert}{\rvert}
\newcommand{\epst}{\epsilon}

\newcommand{\red}[1]{\textcolor{red}{#1}}

\newcommand{\todo}[1]{\textcolor{red}{}}
\newcommand{\note}[1]{\textcolor{blue}{}}

\newcommand{\es}[0]{\mbox{P-SAT}\xspace}
\newcommand{\hes}[0]{\mbox{H-SAT}\xspace}
\newcommand{\at}[0]{AT\xspace}
\newcommand{\trd}[0]{TRADES\xspace}
\newcommand{\dyn}[0]{DAT\xspace}
\newcommand{\cur}[0]{CAT18\xspace}
\newcommand{\cat}[0]{CAT20\xspace}
\newcommand{\iaat}[0]{IAAT\xspace}

\newcommand\scalemath[2]{\scalebox{#1}{\mbox{\ensuremath{\displaystyle #2}}}}

\begin{document}

\fancyhead{}

\title{SAT: Improving Adversarial Training via Curriculum-Based Loss Smoothing}

\author{Chawin Sitawarin}
\affiliation{%
  \institution{University of California, Berkeley}
  \city{Berkeley}
  \state{California}
  \country{USA}
  \postcode{94709}
}
\authornote{Part of the work done while at IBM T. J. Watson Research Center.}
\email{chawins@berkeley.edu}

\author{Supriyo Chakraborty}
\affiliation{%
  \institution{IBM T. J. Watson Research Center}
  \city{Yorktown Heights}
  \state{New York}
  \country{USA}}
\email{supriyo@us.ibm.com}

\author{David Wagner}
\affiliation{%
  \institution{University of California, Berkeley}
  \city{Berkeley}
  \state{California}
  \country{USA}
}
\email{daw@cs.berkeley.edu}

\begin{abstract}
Adversarial training (\at) has become a popular choice for training robust networks. However, it tends to sacrifice clean accuracy heavily in favor of robustness and suffers from a large generalization error. To address these concerns, we propose Smooth Adversarial Training (SAT), guided by our analysis on the eigenspectrum of the loss Hessian. We find that curriculum learning, a scheme that emphasizes on starting ``easy'' and gradually ramping up on the ``difficulty'' of training, smooths the adversarial loss landscape for a suitably chosen difficulty metric. We present a general formulation for curriculum learning in the adversarial setting and propose two difficulty metrics based on the maximal Hessian eigenvalue (\hes) and the softmax probability (\es). 
We demonstrate that SAT stabilizes network training even for a large perturbation norm and allows the network to operate at a better clean accuracy versus robustness trade-off curve compared to \at. This leads to a significant improvement in both clean accuracy and robustness compared to \at, \trd, and other baselines.
To highlight a few results, our best model improves normal and robust accuracy by 6\% and 1\% on CIFAR-100 compared to \at, respectively. On Imagenette, a ten-class subset of ImageNet, our model outperforms \at by 23\% and 3\% on normal and robust accuracy respectively.
\end{abstract}

\ccsdesc[500]{Security and privacy~Domain-specific security and privacy architectures}

\keywords{Adversarial Machine Learning, Adversarial Examples, Curriculum Learning}

\maketitle

\section{Introduction} \label{sec:intro}
It is well-known that machine learning models are easily fooled by adversarial examples, generated by adding carefully crafted perturbation to normal input samples~\cite{szegedy_intriguing_2014,goodfellow_explaining_2015,biggio_evasion_2013}.
This raises serious safety concerns for systems and solutions that rely on machine learning as a crucial component (e.g., identity verification, malware detection, self-driving vehicles). 
Among numerous defenses proposed, Adversarial Training (\at)~\cite{madry_deep_2018} is one of the most widely used algorithm to train neural networks that are robust to adversarial examples.

While the formulation of \at as a robust optimization problem is theoretically sound, solving it for neural networks is indeed tricky.
Here, we focus on two problems from which models trained by \at tend to often suffer. 
First, it tends to sacrifice accuracy on benign samples, by a large margin, to gain robustness or accuracy on adversarial examples. 
This is undesirable for applications that requires high accuracy when operating in the normal settings.
Second, models trained with \at often have a large generalization gap between their train and test adversarial accuracies. 
This gap is typically larger than the generalization error in non-adversarial settings.
In some cases, neural networks are reduced to trivial classifiers that outputs a constant label for all inputs as they fail to learn any better robust decision boundary that actually exists.

Since the objective of \at is sound, we believe that these problems can be mitigated by optimization techniques.
Intuitively, we posit that both the above problems stem from the model being presented with adversarial examples that are ``too difficult to learn from'' at the very beginning of the training which, in turn, causes the model to overfit to such samples.
To this end, we observe that the concept of \emph{curriculum learning}~\cite{bengio_curriculum_2009}, which advocates that model training be initiated with ``easy'' samples before introducing the ``hard'' ones, can naturally help overcome the above problems. 

\begin{figure*}[t]
    \centering
    \hfill
    \begin{subfigure}[b]{0.25\textwidth}
        \centering
        \includegraphics[width=\textwidth]{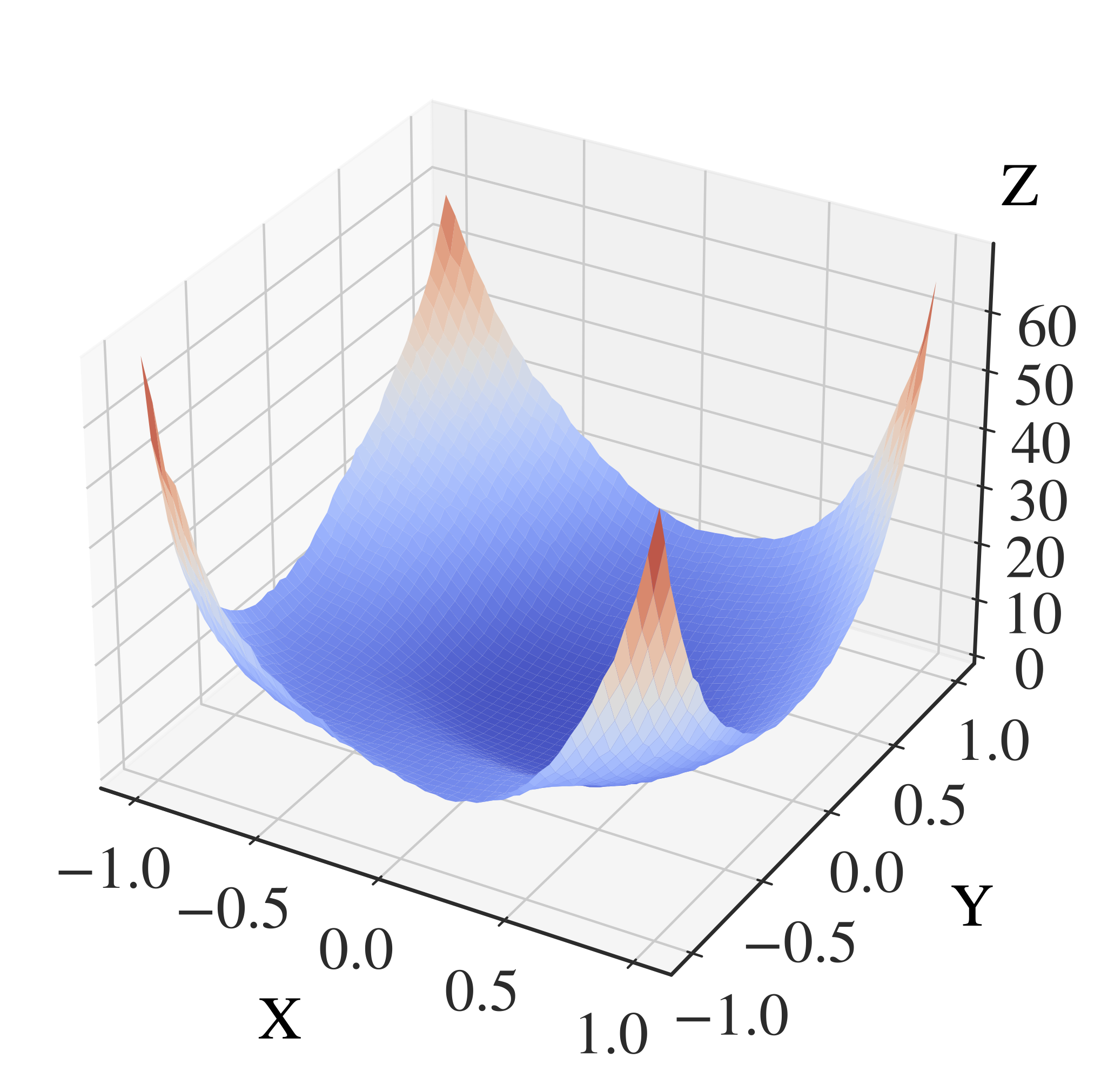}
        \caption{Adversarial Training (\at)}
    \end{subfigure} 
    \hfill
    \begin{subfigure}[b]{0.25\textwidth}
        \centering
        \includegraphics[width=\textwidth]{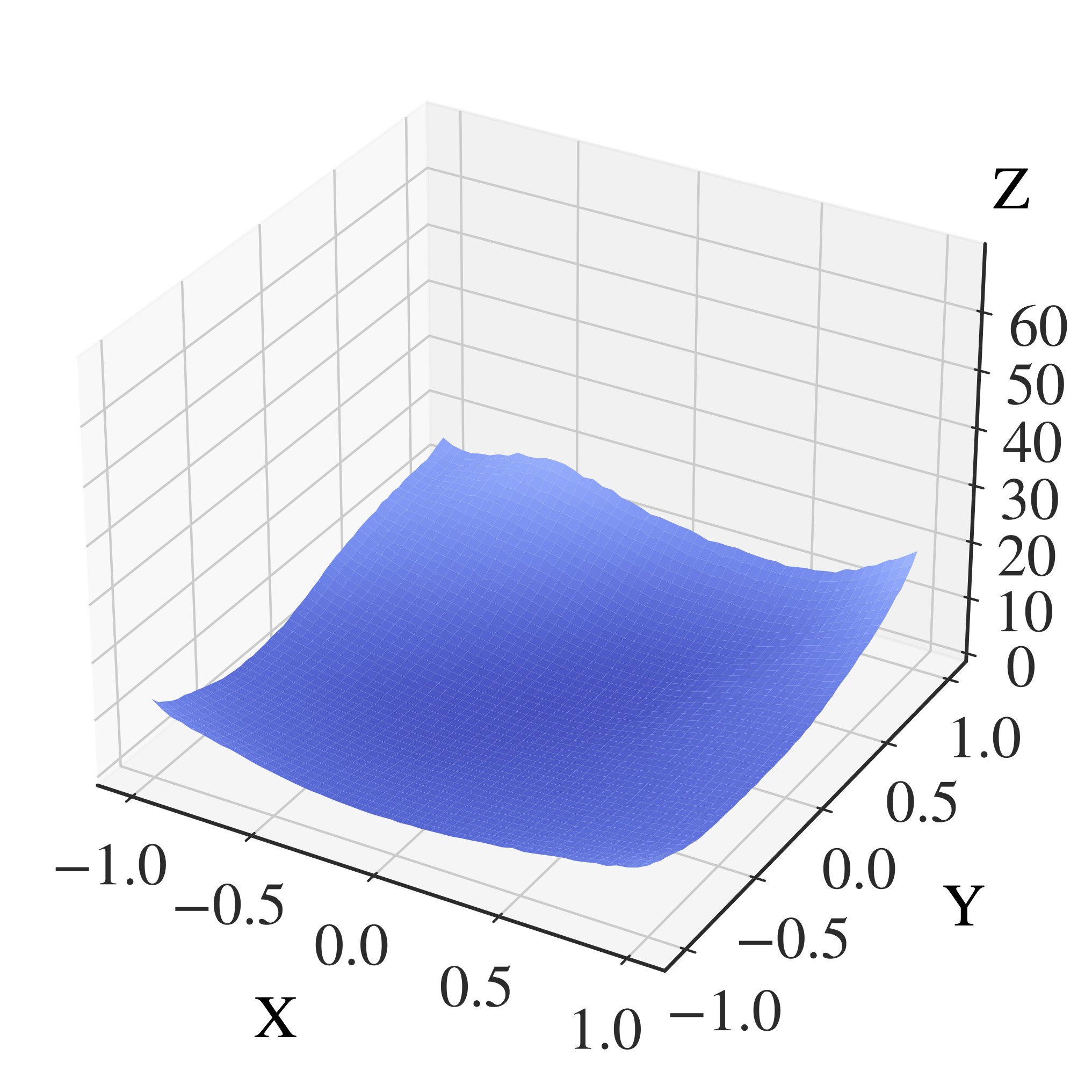}
        \caption{\hes (ours)}
    \end{subfigure}
    \hfill
    \begin{subfigure}[b]{0.25\textwidth}
        \centering
        \includegraphics[width=\textwidth]{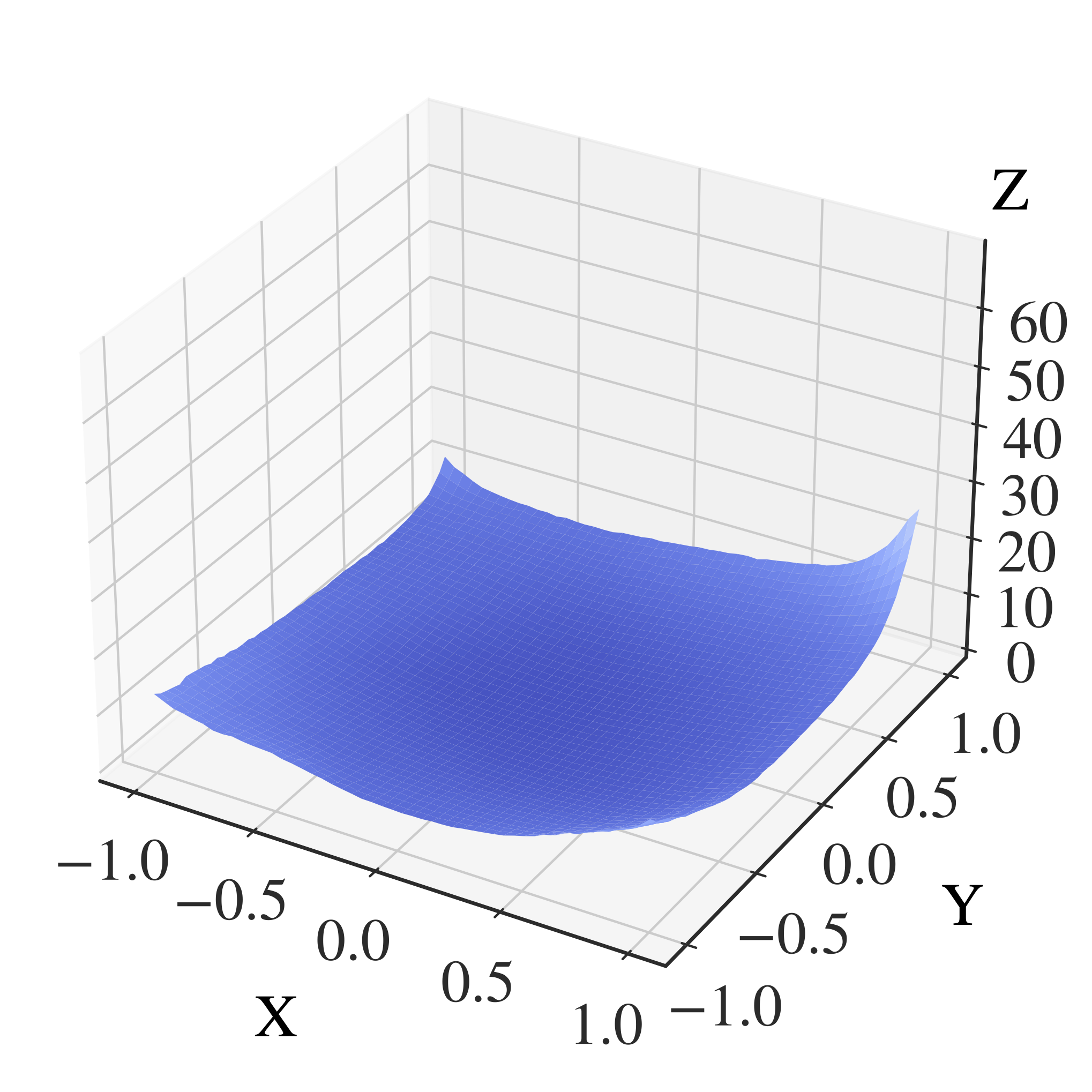}
        \caption{\es (ours)}
    \end{subfigure}
    \hfill
    \caption{Loss landscapes of three PreAct-ResNet-20 models adversarially trained on CIFAR-10 using perturbation $\epsilon=8/255$. The x-axis and y-axis represent two orthogonal random directions in the network parameter space. The z-axis represents the adversarial loss computed with the same normalization proposed by \citet{li_visualizing_2018}. Applying our algorithms, (b) \hes and (c) \es, results in a much smoother loss landscape than (a) \at.}
    \label{fig:loss}
    \vspace{-5pt}
\end{figure*}

Different from prior approaches used in curriculum-based adversarial training algorithms, we draw inspiration from the connection between small generalization error and smooth loss
landscapes~\cite{hochreiter_flat_1997,keskar_largebatch_2017} and design two difficulty metrics---\emph{maximum eigenvalue of the Hessian matrix} and \emph{softmax probability gap}---that directly and indirectly affect the sharpness of the adversarial loss landscape. 
By controlling these metrics, we can bias the network towards flatter regions on the adversarial loss landscape and hence, minimize the generalization gap. 
Following the metrics, we name these two algorithms, \emph{Hessian-Based} Smooth Adversarial Training (\hes) and \emph{Probability-Based} Smooth Adversarial Training (\es).
Following the technique from Li et al.~\cite{li_visualizing_2018}, we visualize the loss landscapes (see Fig.~\ref{fig:loss}) of the networks trained with \at, \hes, and \es to confirm that both of our schemes do lead the networks to smoother regions.

We make the following contributions. First, we unify the prior works on curriculum-based adversarial training under a single formulation. 
Second, we derive techniques to quickly approximate the two difficulty metrics used in \hes and \es so that they can be combined with adversarial training efficiently.
Finally, we systematically compare our proposed method against the baselines using multiple datasets (MNIST, CIFAR-10, CIFAR-100, Imagenette) and network architectures. 
Both our proposed schemes outperform state-of-the-art defenses including TRADES~\cite{zhang_theoretically_2019} and other curriculum-inspired algorithms~\cite{cai_curriculum_2018,wang_convergence_2019,cheng_cat_2020} in term of robustness while maintaining competitive clean accuracy in most settings.
Our best model improves normal and robust accuracy by 6\% and 1\% on CIFAR-100 compared to \at, respectively.
On Imagenette with $\epsilon = 16/255$, our model outperforms \at by 23\% and 3\% on normal and robust accuracy respectively.

\section{Background and Related Work} \label{sec:background}
\paragraph{Adversarial examples.} Adversarial examples are a type of evasion attack against machine learning models generated by adding small perturbations to clean samples~\citep{szegedy_intriguing_2014,goodfellow_explaining_2015,biggio_evasion_2013}. 
The desired perturbation, constrained to be within some $\ell_p$-norm ball, is typically formulated as a solution to the following optimization problem:
\begin{align} \label{eq:adv}
    x^{adv} = x + \delta^* \quad \text{where} \quad \delta^* = ~ &\argmax_{\delta : \norm{\delta}_p \le \epst} ~~ \ell(x + \delta; \theta)
\end{align}
where $\ell(\cdot; \theta):\mathbb{R}^d \to \mathbb{R}$ is the loss function of the target neural network, parameterized by $\theta$, with respect to its input. 
The perturbation is bounded in an $\ell_p$-norm ball of radius $\epst$ and treated as a proxy for imperceptibility of the perturbation. 
Projected gradient descent (PGD) is a popular technique used to solve Eqn.~\eqref{eq:adv}~\citep{madry_deep_2018}.

\paragraph{Defenses against adversarial examples.} In a nutshell, the adversarial training (\at) algorithm, iteratively generates adversarial examples corresponding to every batch of the training data by solving Eqn.~\eqref{eq:adv} and then trains a model by minimizing the expected loss over the adversarial samples. 
\at is formulated as an optimization of the saddle point problem where $\theta$ are the model parameters, and $\{(x_i, y_i)\}_{i=1}^n$ denotes the training set:
\begin{align} \label{eq:at}
    \argmin_\theta ~ &\frac{1}{n} \sum_{i=1}^{n}~ \ell_\epsilon(x_i;\theta) \\
    \text{where} \quad &\ell_\epsilon(x;\theta) \coloneqq \max_{\delta : \norm{\delta}_p \le \epst} ~ \ell(x + \delta; \theta)
\end{align}

Several prior works have attempted to improve \at in terms of computation time~\citep{shafahi_adversarial_2019,wong_fast_2020} and robustness gain~\citep{gowal_uncovering_2021,wu_adversarial_2020}. 
Others have tried different loss functions that are better suited to adversarial training~\citep{zhang_theoretically_2019,ding_mma_2020,wang_improving_2020}. %

\paragraph{Curriculum learning and adversarial training.}
In curriculum adversarial training (\cur)~\citep{cai_curriculum_2018}, the authors create a curriculum by slowly increasing the number of PGD steps during training. But their empirical results suggest that curriculum alone is not effective and it must be combined with other techniques such as quantization and batch mixing. \citet{wang_convergence_2019} (\dyn) defined convergence score, motivated by the Frank-Wolfe optimality gap, as a metric to imitate a curriculum. Two recent works, \citet{balaji_instance_2019} (\iaat) and \citet{cheng_cat_2020} (\cat), use an adaptive and sample-specific perturbation norm during training. Their motivation is that not all samples should be at a fixed distance from the decision boundary as encouraged by \at. Instead, margins should be flexible and data-dependent. In fact, this effect is a by-product of our scheme which perturbs \emph{naturally} easier samples more than harder ones \footnote{Naturally easy samples can be thought of as clean inputs that a given network classifies correctly with high confidence. In other words, easy samples are on the correct side of the decision boundary and are also far from it.}. Concurrent to our work, \citet{zhang_attacks_2020} proposed a formulation of an upper bound of the adversarial loss and, similarly to ours, used early stopping as a realization. However, their criterion is based on the number of PGD steps, while ours, inspired by curriculum learning, relies on a concrete difficulty metric. Furthermore, we also perform extensive evaluation over a number of bench marking datasets to validate our approach.

\paragraph{Smoothness of Loss Landscapes.} 
There is a long-standing hypothesis regarding the correlation between smoothness of the loss surface and the network generalization error~\citep{hochreiter_flat_1997,keskar_largebatch_2017}.
Previous works have successfully improved generalization in both normal and adversarial settings by biasing the training process of the networks towards regions with smooth loss landscapes~\citep{chaudhari_entropysgd_2017,izmailov_averaging_2018,wu_adversarial_2020}.
\emph{In this work, we achieve a similar goal but follow a unique approach via curriculum learning, which has been shown to produce a smoothening effect on the objective~\citep{bengio_curriculum_2009}}.

\section{Adversarial Loss Landscape} \label{sec:loss_landscape}

\subsection{Smoothness and Hessian} \label{ssec:smooth}

Recently, \citet{liu_loss_2020} demonstrated a theoretical correlation between the \emph{sharpness} of the adversarial loss landscape and the perturbation norm $\epsilon$.
A smaller $\epsilon$ implies a smoother loss landscape.
We first briefly restate this result.
In this setup, \citet{liu_loss_2020} assume that the normal loss function, $\ell(x; \theta)$, has Lipschitz continuous gradients w.r.t. $\theta$ and $x$ with constants $L_{\theta \theta}$ and $L_{\theta x}$ respectively.
Mathematically, this can be written as $\forall x, x_1, x_2 \in \mathbb{R}^d$ and $\forall \theta, \theta_1, \theta_2 \in \Omega$,
\begin{align}
    \norm{\nabla_\theta \ell(x; \theta_1) - \nabla_\theta \ell(x; \theta_2)}_2 &\le L_{\theta\theta}\norm{\theta_1 - \theta_2}_2 \\
    \norm{\nabla_\theta \ell(x_1; \theta) - \nabla_\theta \ell(x_2; \theta)}_2 &\le L_{\theta x}\norm{x_1 - x_2}_2
\end{align}
For adversarial loss, $\ell_\epsilon(x; \theta)$, the analogous expression includes an extra term that depends on the perturbation norm $\epsilon$ as follows~\citep{liu_loss_2020}:
\begin{align}
\label{eqn:adv_loss_bound}
    \norm{\nabla_\theta \ell_\epsilon(x;\theta_1) - \nabla_\theta \ell_\epsilon(x;\theta_2)}_2 \le L_{\theta\theta}\norm{\theta_1 - \theta_2}_2 + 2\epsilon L_{\theta x}  
\end{align}

While Eqn.~\eqref{eqn:adv_loss_bound} reveals a relationship between $\epsilon$ and the smoothness of the adversarial loss (LHS of Eqn.~\ref{eqn:adv_loss_bound}), the upper bound is loose and difficult to compute.
This is because Eqn.~\eqref{eqn:adv_loss_bound} holds globally and both $L_{\theta \theta}$ and $L_{\theta x}$ are also global quantities (they hold for all values of $\theta$ and $x$).
Instead, we are only interested in the local smoothness at a particular $\theta$ in practice.

Prior works often quantified smoothness using the eigenspectrum of the Hessian matrix computed locally around a given $\theta$.
To see why it makes sense to use the Hessian, note that we can upper bound the LHS in Eqn.~\eqref{eqn:adv_loss_bound} by the spectral norm of the Hessian matrix using the mean value theorem.
In particular, let $\Delta \theta = \theta_2 - \theta_1$ and $\theta(t) = \theta_1 + t \Delta \theta$, we know that $\exists t \in [0,1]$ s.t.
\begin{align}
    \scalemath{0.99}{\norm{\nabla_\theta \ell_\epsilon(x;\theta_1) - \nabla_\theta \ell_\epsilon(x;\theta_2)}_2 \le \norm{\nabla^2_\theta \ell_\epsilon(x;\theta(t))}_{(2)} \norm{\theta_1 - \theta_2}_2}
\end{align}
We use $\norm{\cdot}_{(2)}$ to denote the spectral norm of a matrix to differentiate it from $\ell_2$-norm of a vector.
Now obviously, if we only consider $\theta_2$ that are close to $\theta_1$, simply choosing $t=0$ yields a good estimate of the local smoothness around $\theta_1$.

Here, we also choose to measure the smoothness locally. 
We use Hessian of the adversarial loss w.r.t. $\theta$, which is defined as Hessian of the normal loss evaluated at the adversarial example for a given $\theta$.\footnote{This should be taken as an approximation rather than the true Hessian for the following reason. Note that the adversarial loss $\ell_\epsilon(x; \theta)$ is a maximal-value function of the normal loss over $x$ (see Eqn.~\eqref{eq:at}). When the normal loss is convex, Danskin's Theorem states that we can use the substitution with $x^*$ to compute first-order derivatives of an extreme-value function. However, here we are considering second-order derivatives of a non-convex function. For a more rigorous analysis, we may need a more generalized version of Danskin's Theorem for Hessian~\citep{shapiro_secondorder_1985}.} 
Specifically, 

\begin{align}
    &\textsc{Smoothness}(\theta) \approx \norm{H_\epsilon(x;\theta)}_{(2)} \\
    &H_\epsilon(x;\theta) \coloneqq \nabla^2_\theta \ell(x^*; \theta) ~ \text{ for } x^* \in \argmax_{z:\norm{z - x}_p \le \epsilon}~ \ell(z;\theta) \label{eqn:smoothness_defn}
\end{align}
Note, the above measure of smoothness, as suggested previously in \citet{liu_loss_2020}, also depend on both $\epsilon$ and the Hessian of $\ell(\cdot)$ w.r.t. $\theta$. 
However, their relationship is differently expressed as in Eqn.~\eqref{eqn:smoothness_defn} compared to that in Eqn.~\eqref{eqn:adv_loss_bound}.
Furthermore, notice that $\norm{H_\epsilon(x,\theta)}_{(2)}$ is the absolute value of the largest eigenvalue of $H_\epsilon(x,\theta)$ since any Hessian matrix is symmetric. 
We will refer to this quantity as ``maximal Hessian eigenvalue.''

\subsection{Curriculum Learning and Smoothness} \label{ssec:cl_smooth}

We believe that smooth loss landscapes will benefit the adversarial training process in two ways: generalization and convergence.
First, as mentioned in Section~\ref{sec:background}, many previous works have shown connections between generalization and smoothness of the loss landscape.
Flat minima introduce an \emph{implicit bias} for SGD on deep neural networks and has served as an explanation to the surprisingly good generalization of over-parameterized neural networks.
Examples of factors that affect this implicit bias include batch size, learning rate, and architecture such as residual connections~\citep{keskar_largebatch_2017,jastrzebski_three_2018,li_visualizing_2018,mulayoff_unique_2020}.
Other works also aim to ``artificially'' create such flat local minima through more advanced training techniques~\citep{chaudhari_entropysgd_2017,izmailov_averaging_2018}.

The second benefit of a smooth loss surface is a faster convergence. 
It is well-known that smoother loss (e.g., smaller Lipschitz gradient constant) allows for a large step size and hence, a faster convergence on both convex and non-convex problems including ERM on neural networks~\citep{lee_gradient_2016,khamaru_convergence_2019}.
Conversely, a sharp loss surface is harmful to the training process.
\citet{liu_loss_2020} show that a large adversarial perturbation could increase the sharpness, i.e., make gradients large even near the minima, and hence, slow down the training.

In this work, we explore curriculum learning as a mechanism to encourage smoothness of the adversarial loss surface.
The original intuition behind curriculum learning as outlined by \citet{bengio_curriculum_2009} was to help smooth the loss landscape in the initial phase of non-convex optimization.
While this intuition has found acceptance in the community, to the best of our knowledge, this notion of smoothness has not been used for curriculum learning.
Additionally, curriculum learning is particularly suitable in the adversarial context because it can, directly or indirectly, manipulate $\epsilon$, which is a main factor affecting the smoothness as mentioned previously (Section~\ref{ssec:smooth}).
In Section~\ref{sec:ates}, we make this connection explicit via our proposed algorithms.

\section{Smoothed Adversarial Training} \label{sec:ates}

\subsection{Curriculum Learning and Difficulty Metric}

It has been both empirically and theoretically (for a linear regression model) established that curriculum learning improves the early convergence rate as well as the final generalization performance, especially when the task is difficult (e.g., under-parameterized models, heavy regularization)~\citep{bengio_curriculum_2009,weinshall_curriculum_2018,hacohen_power_2019}.
The main challenge therefore lies in defining an effective difficulty metric to dictate the order of the training samples presented to the network.
Luckily, the adversarial setting provides several intuitive difficulty metrics that can be easily controlled through the perturbation norm, $\epsilon$. 

We start by proposing a formulation of the curriculum-augmented adversarial loss, or \emph{curriculum loss}, denoted by $\ell_{\psi, \epsilon}$ where $\psi:\mathbb{R}^d \to \mathbb{R}$ is a given difficulty metric acting as an additional constraint on the adversarial loss. We call it the \textit{curriculum constraint}.
\begin{align} \label{eq:cl}
    \ell_{\psi, \epsilon}(x, \lambda) \quad = \quad \max_{\delta:\norm{\delta}_\infty \le \epst} \quad &\ell(x+\delta) \\
    \text{s.t.} \quad &\psi(x+\delta) \le \lambda \nonumber
\end{align}
This general formulation above unifies prior works on curriculum-inspired adversarial training (see Table~\ref{tab:metrics} in Appendix~\ref{ap:sec:constraint}). 
Without the curriculum constraint, the curriculum loss reduces to the normal adversarial loss in Eqn.~\eqref{eq:adv}. 
The difficulty parameter $\lambda$ should be scheduled to increase as the training progresses such that it reaches its maximal value well before the end of training.
This is to ensure that the curriculum loss converges to the original adversarial loss.

As mentioned in Section~\ref{ssec:cl_smooth}, we propose two difficulty metrics that directly aim to smoothen the loss landscape: \emph{Maximal Eigenvalue of the Hessian} and \emph{Softmax Probability Gap}.

\subsection{Maximal Eigenvalue of the Hessian} \label{ssec:hes}

To encourage smoothness, we can directly control the largest eigenvalue of the Hessian throughout the training.
In Section~\ref{ssec:smooth}, we described the correlation between the largest eigenvalue of the Hessian and adversarial strength and explained its usage as a curriculum constraint.
We now use it as a difficulty metric,
\begin{align}
    \psi_H(x) \coloneqq \norm{H_\epsilon(x;\theta)}_{(2)}
\end{align}
for our \emph{Hessian-Based Smooth Adversarial Training} (\hes). 

\begin{figure}[!ht]
    \centering
    \hfill
    \begin{subfigure}[b]{0.2\textwidth}
        \centering
        \includegraphics[width=\textwidth]{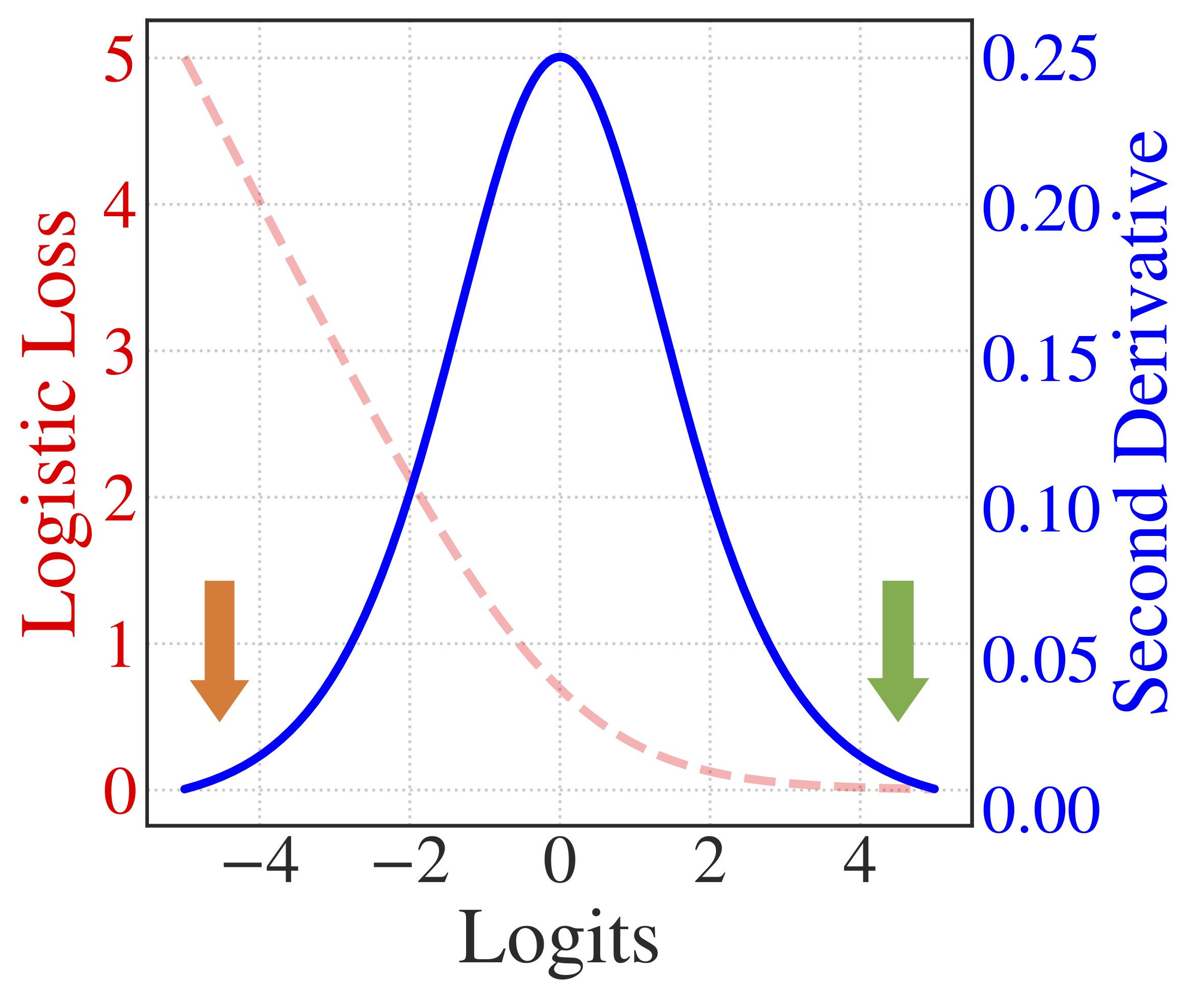}
        \caption{Second derivatives}
    \end{subfigure} 
    \hfill
    \begin{subfigure}[b]{0.2\textwidth}
        \centering
        \includegraphics[width=\textwidth]{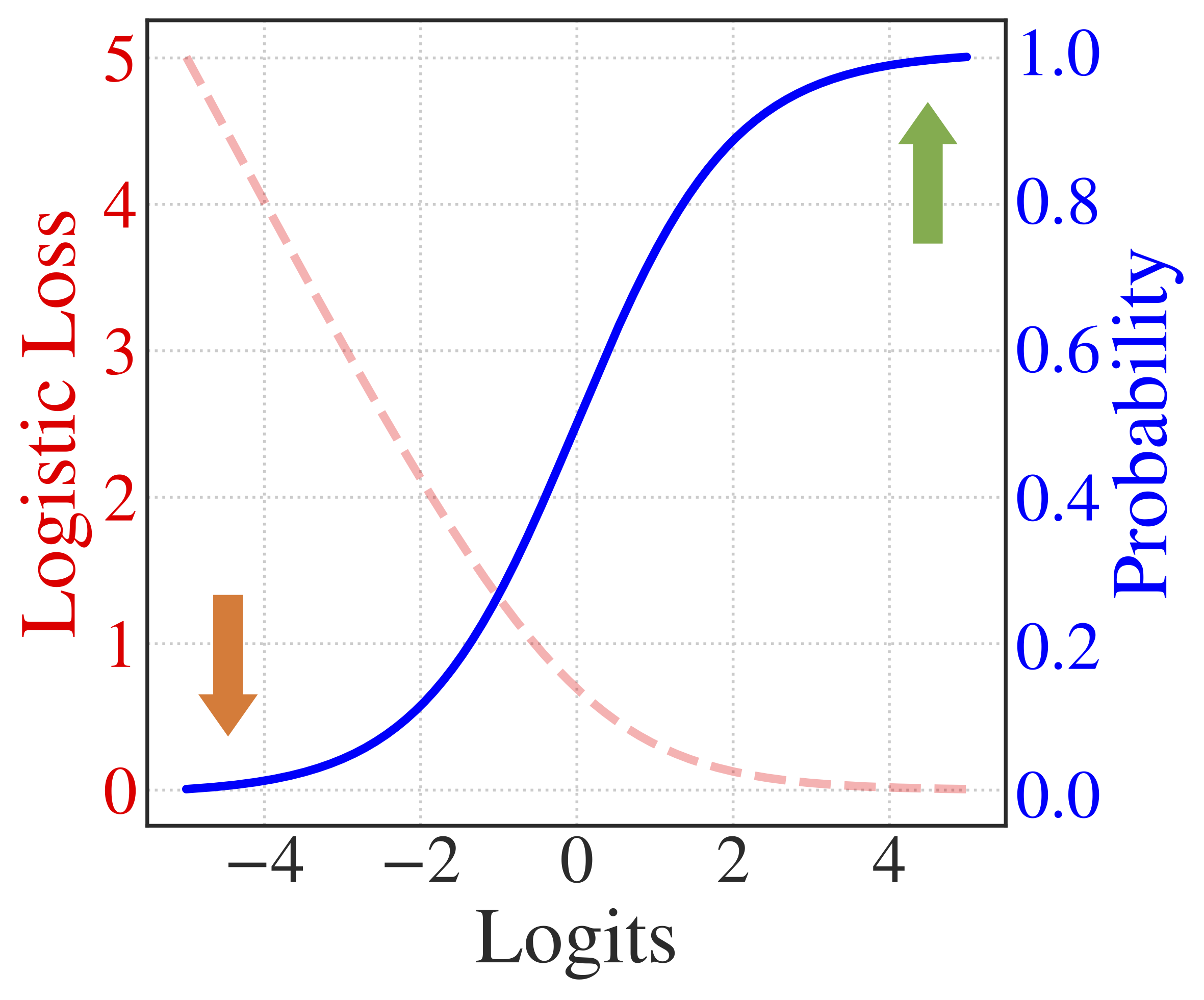}
        \caption{True-class probability}
    \end{subfigure}
    \hfill\phantom{~}
    \caption{Relationship between logistic loss, (a) its second order derivative w.r.t. logit, and (b) the true-class (positive) probability. Small second derivative can correspond to difficult inputs (i.e., high loss, small logit, small probability, orange arrow) which is undesirable.}
    \label{fig:logistic}
    \vspace{-5pt}
\end{figure}

This direct control on the Hessian, however, has two limitations. 
First, computing $\psi_H(x)$ requires calculating the maximal eigenvalue of the Hessian which is very expensive to execute at every PGD step or even training step.
To mitigate this problem, we devise several approximations which significantly speed up the calculation (see Section~\ref{sec:compute}).
Second, the maximal Hessian eigenvalue is small not only when the input is easy but also when it is hard, i.e., when the probability of the correct class is close to either $1$ (for easy) or $0$ (for difficult). 
This might lead to an undesirable side effect where keeping the maximum Hessian eigenvalue small does not guarantee that only easy samples are presented in the early phases of training.
We illustrates this effect for logistic regression in Fig.~\ref{fig:logistic}a.
Note, the second derivative value is small both when the logit value is large and when it is small (a large negative number), corresponding to the green and orange arrows respectively.

As shown in Fig.~\ref{fig:logistic}b, one way to circumvent this limitation of \hes is to directly control the class probability of a sample instead.
If the class probabilities are large, it ensures that the samples are easy, and it is likely that the second derivative (and hence, the maximal Hessian eigenvalue) is also small. 
Since class probability for a sample is available after every PGD step, we will also solve the first problem simultaneously. 
We will explore this option for the difficulty metric in the next section below.

\subsection{Softmax Probability Gap} 

To overcome some of the limitations of \hes, we propose \emph{softmax probability gap}---difference between the true class probability and the largest softmax probability excluding the true class---as the second difficulty metric for our scheme. 
Formally, it is defined as
\begin{align}
    \psi_P(x) \coloneqq \max_{j \ne y} f(x)_j - f(x)_y
\end{align}
where $y \in \{1,...,c\}$ is the ground-truth label of $x$, and $f:\mathbb{R}^d \to \mathbb{R}^c$ is the softmax output of a neural network. 

The probability gap has an intuitive interpretation and is bounded between $-1$ and $1$.
A perturbed input with a large gap, i.e., $\psi_P(x) \approx 1$, means that it has been incorrectly classified with high confidence by the network, suggesting that it is a ``hard'' sample. 
On the other hand, if $\psi_P(x) \approx -1$, the input must be ``easy'' since the network has classified it correctly with high confidence. 
When the gap is zero, the input is right on the decision boundary of the classifier.

\begin{figure}[t]
\centering
\includegraphics[width=0.45\textwidth]{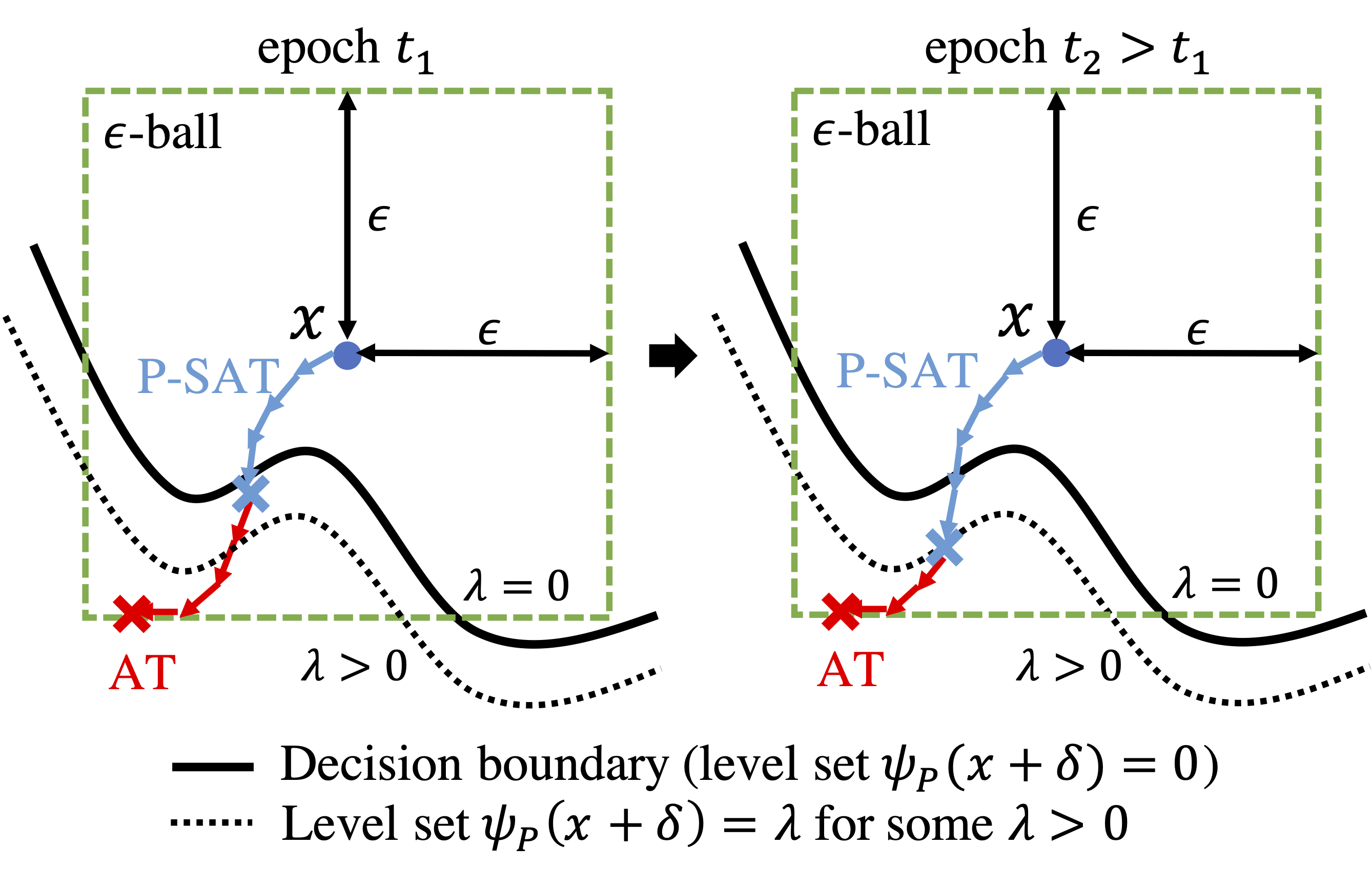}
\vspace{-5pt}
\caption{Comparison between the perturbed samples generated by \at and \es. The blue and red arrows represents PGD updates. For \es, the process stops when the curriculum constraint is violated (blue cross). As $\lambda$ increases, \es approaches \at.}
\label{fig:ates_diag}
\vspace{-5pt}
\end{figure}

When used as a curriculum constraint $\psi_P(x) \le \lambda$, softmax probability gap is also directly related to traditional adversarial training.
When $\lambda = 1$, the constraint is always satisfied for any $x$ and $\delta$. 
Thus, the curriculum loss reduces to the normal adversarial loss. 
The curriculum loss is also a lower bound of the adversarial loss for any $\lambda$ with equality for $\lambda \ge 1$.
\begin{align}
  &\forall\lambda, ~\ell_{\psi, \epsilon}(x, \lambda) \le \ell_\epsilon(x) \quad \text{and} \\
  &\forall\lambda \ge 1, ~\ell_{\psi, \epsilon}(x, \lambda) = \ell_\epsilon(x)
\end{align}
With the above interpretation in mind, we create a curriculum for adversarial training by progressively increasing $\lambda$ from $0$ to $1$ during the training (illustrated in Fig.~\ref{fig:ates_diag}).
First, we expose the model to a weak adversary, or equivalently an easy objective, in the early stage of the training (small $\lambda$).
Progressively, the curriculum objective becomes more difficulty (large $\lambda$) and eventually reach the original adversarial loss when $\lambda=1$.
This algorithm is named \emph{Probability-Based Smooth Adversarial Training} (\es).

\begin{algorithm}[t]
    \SetKwInOut{Input}{Input}
    \SetKwInOut{Output}{Output}
    \SetKwInOut{Parameters}{Parameters}
    \Input{Training set $(X, Y)$, neural network weights $\theta$}
    \Parameters{PGD steps $K$, step size $\eta$, difficulty scores $\{\lambda^t\}_{t=1}^T$, projection $\Pi$ as a function of $p$ and $\epsilon$}
    \For{t = 1,\dots,T}{
        Sample $\{(x_i,y_i)\}_{i=1}^B$ from $(X, Y)$. \\
        Initialize all $\delta_i$'s at random. \\
        \textcolor{blue}{\# Generating adversarial examples with curriculum} \\
        \For{k = 1,\dots,K}{
            \textcolor{blue}{\# Mask out updates that violate the constraint} \\
            $mask_i = \bm{1}\left\{\psi(x_i) \le \lambda^t\right\}$ \\
            \textcolor{blue}{\# Projection step} \\
            $\delta_i \leftarrow \Pi(x_i + \eta \nabla_x \ell(x_i)) - x_i$ \\
            \textcolor{blue}{\# Update samples with the given mask} \\
            $x_i \leftarrow x_i + mask_i * \delta_i$
        }
        \textcolor{blue}{\# Network training step} \\
        Compute $\ell(x_i)$ and update weights $\theta^t$.
    }
    \caption{\es}
    \label{alg}
\end{algorithm}

\subsection{Early Termination of PGD}

Both the curriculum constraints are non-convex and thus it is difficult to solve Eqn.~\eqref{eq:cl} directly as a constrained optimization problem. 
Conceptually, we can satisfy the constraint by terminating PGD as soon as the constraint is violated. 
This is a heuristic for most choices of the difficulty metric $\psi$, but in the case of \es, the early termination solves the optimization exactly for the binary class case as stated in Proposition~\ref{prop:early} below.
For \hes, the early termination is slightly more complicated and is described in Appendix~\ref{ap:ssec:hes_early_stop}.

\begin{restatable}[]{prop}{earlystop}
\label{prop:early}
    In a binary-class problem, an optimum of the curriculum loss with $\psi_P(\cdot)$ as the curriculum constraint can be found by a projected gradient method that terminates when the curriculum constraint is violated.
\end{restatable}

Below we will explain the intuition behind this proposition and defer the complete derivation to Appendix~\ref{ap:ssec:es_early_stop} for improved readability.
Note, the loss function (negative log-likelihood) is a monotonic function of the output class probability.
This allows us to simplify the curriculum loss in Eqn.~\eqref{eq:cl} as follows:
\begin{align}
    \scalemath{1.}{\ell_{\psi,\epsilon}(x, \lambda)  =  \max_{\delta:\norm{\delta}_p \le \epst} \min\{\ell(x+\delta), -\log\left((1-\lambda)/2\right)\}}
    \label{eq:sketch}
\end{align}
The second term of the piecewise minimization in Eqn.~\eqref{eq:sketch} is just a constant. 
This modified problem can be solved with PGD, similarly to the adversarial loss, and terminated as soon as the first term is larger than the second---which is equivalent to the curriculum constraint being violated. 
The curriculum loss is slightly more complicated for the multi-class case, but we propose to solve it approximately using the same early termination as a heuristic. 
Algorithm~\ref{alg} summarizes the implementation of \es. 

\section{Computing Maximal Hessian Eigenvalues} \label{sec:compute}

In this section, we detail techniques to compute the maximal Hessian eigenvalue for $\psi_H(\cdot)$ efficiently.
Computing the full Hessian matrix for neural networks with millions of parameters is computationally expensive and, in fact, unnecessary. 
We apply the power method~\citep{lecun_automatic_1992} to compute the top eigenvalues of the Hessian directly. 
The method generally converges in a few iterations each of which requires two backward passes.

However, the power method is still too expensive to run for every PGD step and training iteration of the already computation-intensive adversarial training.
Therefore, to significantly minimize the computational overhead, we approximate the maximal Hessian eigenvalue using the second-order Taylor's expansion,
\begin{align}
    \ell(\theta + \zeta) = \ell(\theta) + \zeta^\top \nabla_\theta \ell(\theta) + \frac{1}{2}\zeta^\top \nabla^2_\theta \ell(\theta) \zeta + \mathcal{O}(\norm{\zeta}_2^3)
\end{align}
combined with the fact that 
\begin{align}
    \frac{1}{2} \norm{\nabla^2_\theta \ell(\theta)}_{(2)} = \max_{\norm{\zeta}_2 = 1} \frac{1}{2}\abs{\zeta^\top \nabla^2_\theta \ell(\theta) \zeta}.
\end{align}
As we are only interested in the largest positive eigenvalue, we can ignore the absolute value sign.\footnote{Assuming that we are near local minima, most (if not all) eigenvalues should be positive with only a few very small negative eigenvalues.}
Letting $g \coloneqq \nabla_\theta \ell(\theta) / \norm{\nabla_\theta \ell(\theta)}_2$, we can now approximate the upper bound ($\overline{\sigma_1}$) and the lower bound ($\underline{\sigma_1}$) of the maximal eigenvalue of the Hessian ($\sigma_1$) by appropriately setting $\zeta = \pm \alpha g$ for some small constant $\alpha$:
\begin{align}
    &\underline{\sigma_1} ~\lessapprox~ \norm{\nabla^2_\theta \ell(\theta)}_{(2)} ~\lessapprox~ \overline{\sigma_1} \\
    \underline{\sigma_1} ~=~ &\frac{1}{\alpha}\max \{ \ell(\theta + \alpha g) - \alpha\norm{\nabla_\theta \ell(\theta)}_2, \label{eq:lb1} \\ 
    &~\quad\quad\quad\ell(\theta - \alpha g) + \alpha\norm{\nabla_\theta \ell(\theta)}_2 \} - \ell(\theta) \nonumber \\
    \overline{\sigma_1} ~=~ &\frac{1}{\alpha}\left\{\ell(\theta + \alpha g) + \alpha\norm{\nabla_\theta \ell(\theta)}_2 \right\}- \ell(\theta) \label{eq:ub1}
\end{align}
Again, for improved readability, the full derivation and the implementation considerations are provided in Appendix~\ref{ap:ssec:hess_approx}.

Evaluating Eqn.~\eqref{eq:ub1} and \eqref{eq:lb1} at the adversarial example of $x$, gives us the upper and a lower bound respectively, for $\norm{H_\epsilon(x,\theta)}_{(2)}$.
It is also important to note that setting $\alpha$ sufficiently small makes our approximation more accurate in practice. 
In fact, we empirically validated that the lower bound is indeed very tight (within 15\% from the true value).

\section{Experiments} \label{sec:experiment}
\subsection{Setup}
We train and test the robustness of our proposed schemes as well as the baselines on four image datasets, namely MNIST, CIFAR-10, CIFAR-100, and Imagenette~\citep{howard_fastai_2021}, a more realistic and higher-dimensional dataset containing a 10-class subset of the full ImageNet samples.
We use a small CNN for MNIST, Pre-activation ResNet-20 (PRN-20) and WideResNet-34-10 (WRN-34-10) for CIFAR-10/100, and ResNet-34 for Imagenette.
For evaluation, we use AutoAttack~\citep{croce_reliable_2020}, a novel attack based on an ensemble of four different attacks that are collectively stronger than PGD and capable of avoiding gradient obfuscation issue~\citep{athalye_obfuscated_2018}. 
We compare our \hes and \es to four baselines: \at, \dyn, \cat, and \trd.\footnote{We do not include the results on \citet{cai_curriculum_2018} because $\epsilon$-scheduling has been shown to perform worse than \dyn. The scheme uses a non-differentiable component which makes evaluation more complicated and not comparable to the other baselines. We also found that it suffers from gradient obfuscation~\citep{athalye_obfuscated_2018}.}
For more details on the setup, see Appendix~\ref{ap:sec:setup}.

\subsection{Results}
 
First, we compare the defenses in terms of their robustness, accuracy, and a combination of the two.
We report the sum of the clean and the adversarial accuracy as a single metric to represent the robustness-accuracy trade-off.
We do not argue that it is the only correct (or the best) metric but rather a simple one among many other options (e.g., a weighted sum or a non-linear function).
We highlight higher performance gains of our schemes for more difficult tasks and show that they guide the networks towards smoother loss landscapes and improved local minima compared to \at.

\begin{table*}[t]
\small
\centering
\caption{Clean and adversarial accuracy of the defenses on two models (PreAct-ResNet-20 and WideResNet-34-10) for the CIFAR-10 and CIFAR-100 datasets. ``Sum'' indicates the sum of the clean and the adversarial accuracy. The largest number in each column is shown in bold.}
\label{tab:cifar}
\begin{tabular}{@{}lrrrrrrrrrrrr@{}}
\toprule
\multirow{2}{*}{Defenses} & \multicolumn{3}{c}{CIFAR-10 (PRN-20)} & \multicolumn{3}{c}{CIFAR-10 (WRN-34-10)} & \multicolumn{3}{c}{CIFAR-100 (PRN-20)} & \multicolumn{3}{c}{CIFAR-100 (WRN-34-10)}\\ 
\cmidrule(l){2-4} \cmidrule(l){5-7} \cmidrule(l){8-10} \cmidrule(l){11-13}
 & Clean & Adv & Sum & Clean & Adv & Sum & Clean & Adv & Sum & Clean & Adv & Sum \\ \midrule
\at~\cite{madry_deep_2018} & 80.67	& 45.19	& 125.86 & 86.18 & 49.72 & 135.90 & 51.76 & 21.93 & 73.69 & 60.77 & 24.54 & 85.31 \\
\trd~\cite{zhan_theoreticallygrounded_2016} & 80.50 & \textbf{45.77} & 126.27 & 88.08 & 45.83 & 133.91 & 54.73 & 20.17 & 74.90 & 58.27 & 23.57 & 81.84 \\
\dyn~\cite{wang_convergence_2019} & 81.83 & 42.97 & 124.80 & 86.72 & 45.38 & 132.10 & 54.65 & 20.85 & 75.50 & 54.71 & 20.35 & 75.06 \\
\cat~\cite{cheng_cat_2020} & \textbf{86.46} & 21.69 & 108.15 & \textbf{89.61} & 34.78 & 124.39 & 49.29 & 13.13 & 62.42 & 62.84 & 16.82 & 79.66 \\ \midrule
\hes (ours) & 81.85 & 44.88 & 126.73 & 85.56 & 47.25  & 132.81 & 56.64 & 22.25 & 78.89 & 61.33 & \textbf{25.43} & 86.76 \\
\es (ours)  & 83.99 & 44.54 & \textbf{128.53} & 86.84 & \textbf{50.75} & \textbf{137.59} & \textbf{57.90} & \textbf{22.93} & \textbf{80.83} & \textbf{62.95} & 24.56 & \textbf{87.51} \\ 
\bottomrule  
\end{tabular}
\end{table*}

\begin{table}[t]
\small
\centering
\caption{Clean and adversarial accuracy of the defenses on MNIST dataset. Adversarial accuracy is measured by AutoAttack for $\epsilon=0.3$ and $\epsilon=0.45$. The numbers in red indicate that the network is stuck in a sub-optimal local minimum.}
\label{tab:mnist}
\begin{tabular}{@{}lrrrrrr@{}}
\toprule
\multirow{2}{*}{Defenses}  & \multicolumn{3}{c}{$\epsilon = 0.3$}  & \multicolumn{3}{c}{$\epsilon = 0.45$}  \\ \cmidrule(l){2-4} \cmidrule(l){5-7} 
& Clean & Adv & Sum & Clean & Adv & Sum \\ \midrule
\at~\citep{madry_deep_2018} & 98.07 & 85.47 & 183.54 & \red{11.22} & \red{11.22} & \red{22.44} \\
\trd~\citep{zhan_theoreticallygrounded_2016} & 98.98 & 90.70 & 189.68 & 97.36 & 0.00 & 97.36 \\
\dyn~\citep{wang_convergence_2019} & 98.93 & \textbf{92.24} & \textbf{191.17} & 97.98 & \textbf{65.71} & \textbf{163.69} \\
\cat~\citep{cheng_cat_2020} & \textbf{99.46} & 0.00 & 99.46 & \textbf{99.39} & 0.00 & 99.39 \\ \midrule
\hes (ours) & 99.01 & 80.71 & 179.72 & 98.35 & 54.10 & 152.45 \\
\es (ours) & 99.16 & 92.00 & 191.16 & 97.87 & 58.50 & 156.37 \\ 
\bottomrule  
\end{tabular}
\vspace{-5pt}
\end{table}

\subsubsection{Comparing clean and adversarial accuracy} 

On MNIST (shown in Table~\ref{tab:mnist}), with $\epsilon$ of $0.3$, both the clean and the adversarial accuracy, fall in the same range, for all defenses except for \cat.
Our \es has very similar accuracies to \dyn and outperforms the rest of the baselines.
\cat experiences gradient obfuscation~\citep{athalye_obfuscated_2018}, resulting in an over-estimated adversarial accuracy against PGD attack.
However, the true robustness is very low as revealed by the stronger AutoAttack.
This issue persists on \cat for all the datasets we tested and is likely caused by their label smoothing.

For $\epsilon=0.45$, the difference becomes significant. 
Apart from \hes, \es, and \dyn, none of the other defenses are robust, having adversarial accuracy of ${\sim}10\%$ or lower. 
The clean and the adversarial accuracy of \at are the same and are close to a random guess because it outputs the same logit values for every input. 
This phenomenon happens when the training gets stuck in a sub-optimal local minimum where the network ``finds an easy way out'' and resorts to learning a trivial solution. 
Changing the optimizer, the learning rate, or the random seed can sometimes mitigate the problem, but we found that no combination of these changes allowed \at to effectively train on MNIST with $\epsilon \ge 0.35$.
It is likely caused by the non-smooth loss landscape which is amplified by a large value of $\epsilon$.
Importantly, the fact that only curriculum-based adversarial training methods (\hes, \es, and \dyn) works in this case suggests that they help smoothen the loss surface.

In Table~\ref{tab:cifar}, our scheme is more robust than all other defenses on both PRN-20 and WRN-34-10 for CIFAR-10 and CIFAR-100 datasets with an exception of PRN-20 on CIFAR-10 where TRADES is the most robust.
\es also has the highest sum of clean and adversarial accuracy in all cases.
\cat generally has slightly higher clean accuracy than the other models, but it is the least robust by a large margin due to the false robustness issue previously mentioned.
We note that on most settings, \at is a very strong baseline. 
It has roughly the same performance as \trd and outperforms the other curriculum-based schemes in many experiments. 
This may seem surprising, but it is also observed by \citet{rice_overfitting_2020} and \citet{gowal_uncovering_2021} that when the training is stopped early to prevent overfitting (as in our experiments), \at performs as well as other more complicated defenses.

As expected, \es outperforms \hes in most settings.
This is likely due to the two limitations we mentioned in Section~\ref{ssec:hes}: (1) additional approximations introduced to \hes for practical consideration as well as (2) the fact that the Hessian eigenvalue does not always correspond to difficulty.
While \es controls the smoothness in a less direct manner, it does not suffer from these two issues. 
Consequently, it is computationally cheaper and also performs slightly better.

\begin{table*}[t]
\small
\centering
\caption{Clean and adversarial accuracy of the defenses on CIFAR-10 and Imagenette datasets trained on PreAct-ResNet-20 and ResNet-34 respectively.  Here, we use a larger perturbation norm $\epsilon$ of $16/255$ and $24/255$.}
\label{tab:cifar-img}
\begin{tabular}{@{}lrrrrrrrrrrrr@{}}
\toprule
\multirow{3}{*}{Defenses} & \multicolumn{6}{c}{CIFAR-10} & \multicolumn{6}{c}{Imagenette} \\
\cmidrule(l){2-7} \cmidrule(l){8-13} 
& \multicolumn{3}{c}{$\epsilon = 16/255$} & \multicolumn{3}{c}{$\epsilon = 24/255$} & \multicolumn{3}{c}{$\epsilon = 16/255$} & \multicolumn{3}{c}{$\epsilon = 24/255$} \\ 
\cmidrule(l){2-4} \cmidrule(l){5-7} \cmidrule(l){8-10} \cmidrule(l){11-13}
& Clean & Adv & Sum & Clean & Adv & Sum & Clean & Adv & Sum & Clean & Adv & Sum \\ \midrule
\at~\cite{madry_deep_2018} & 63.66	& \textbf{23.78} & 87.44 & 41.68 & \textbf{15.21} & 56.89 & 49.10	& 28.00 & 77.10 & 42.55	& 21.05 & 63.60 \\
\trd~\cite{zhan_theoreticallygrounded_2016} & \textbf{68.70} & 17.20 & 85.90 & \textbf{58.83} & 5.74 & 64.57 & \textbf{78.05} & 8.90 & 86.95 & \textbf{68.50} & 1.90 & 70.40 \\
\dyn~\cite{wang_convergence_2019} & 57.28 & 18.94 & 76.22 & 31.21 & 14.36 & 45.57 & 66.20 & 30.30 & 96.50 & 52.50 & 24.50 & 77.00 \\ \midrule
\hes (ours) & 64.27 & 23.19 & 87.46 & 49.85 & 13.35 & 63.20 & 69.10 & \textbf{35.45} & \textbf{104.55} & 47.50 & \textbf{27.75} & 75.25 \\
\es (ours) & 66.99 & 22.42 & \textbf{89.41} & 52.58 & 12.24 & \textbf{64.82} & 72.20 & 31.25 & 103.45 & 62.15	& 20.00 & \textbf{82.15} \\ \bottomrule  	
\end{tabular}
\end{table*}

\begin{figure*}[ht!]
    \centering
    \hfill
    \begin{subfigure}[b]{0.3\textwidth}
        \centering
        \includegraphics[width=\textwidth]{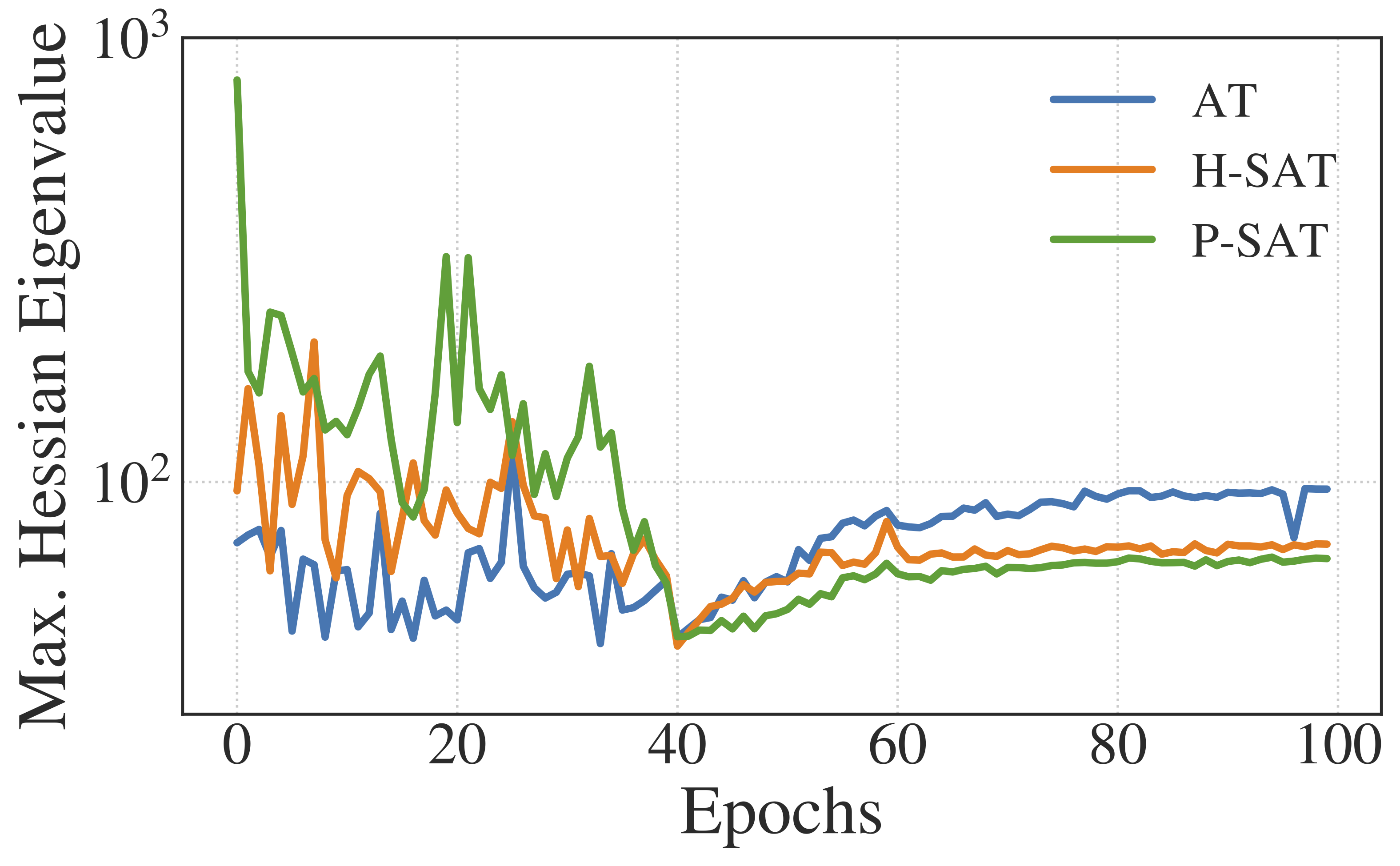}
        \caption{Maximal Hessian eigenvalue}
    \end{subfigure} 
    \hfill
    \begin{subfigure}[b]{0.29\textwidth}
        \centering
        \includegraphics[width=\textwidth]{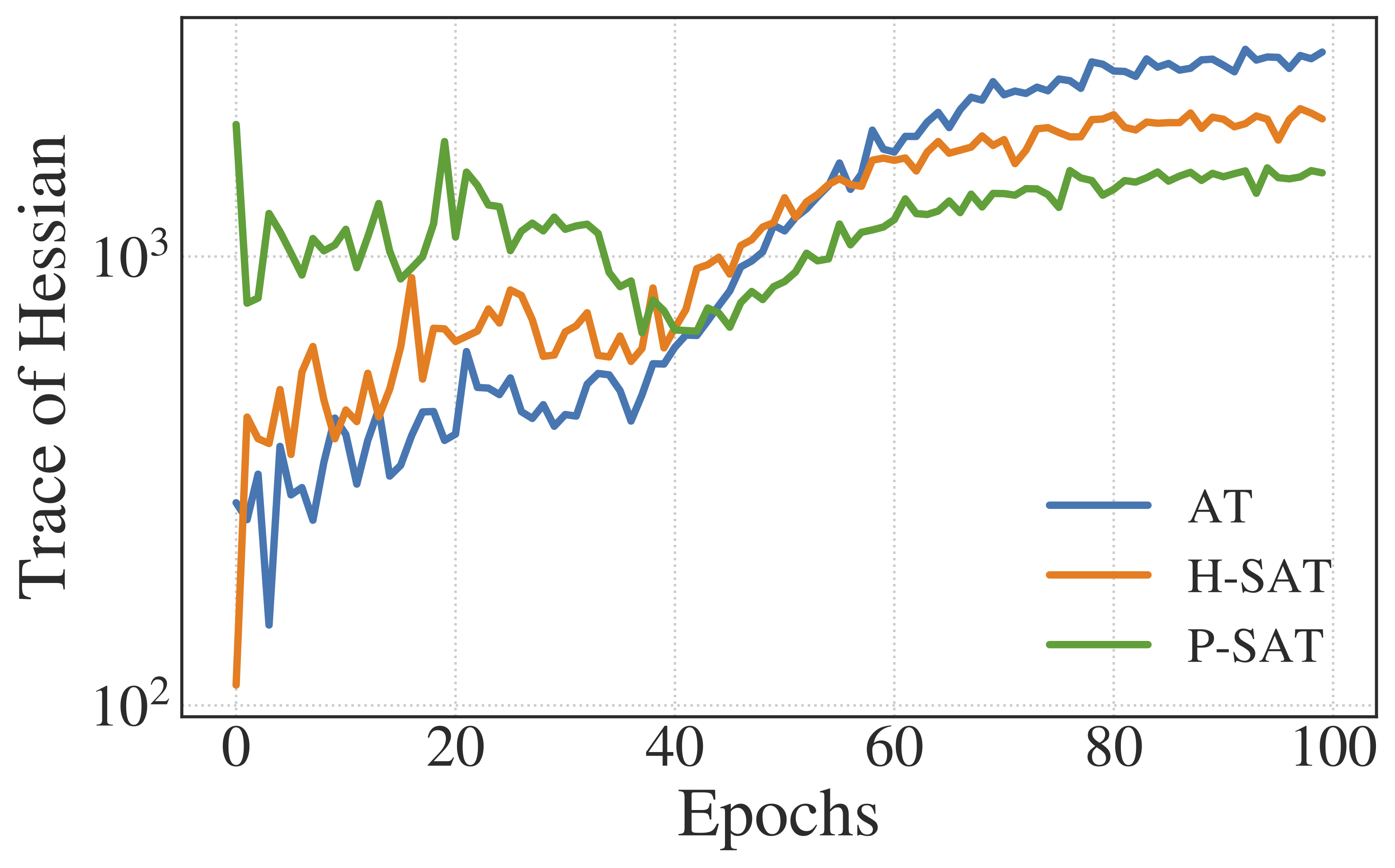}
        \caption{Trace of Hessian}
    \end{subfigure}
    \hfill
    \begin{subfigure}[b]{0.29\textwidth}
        \centering
        \includegraphics[width=\textwidth]{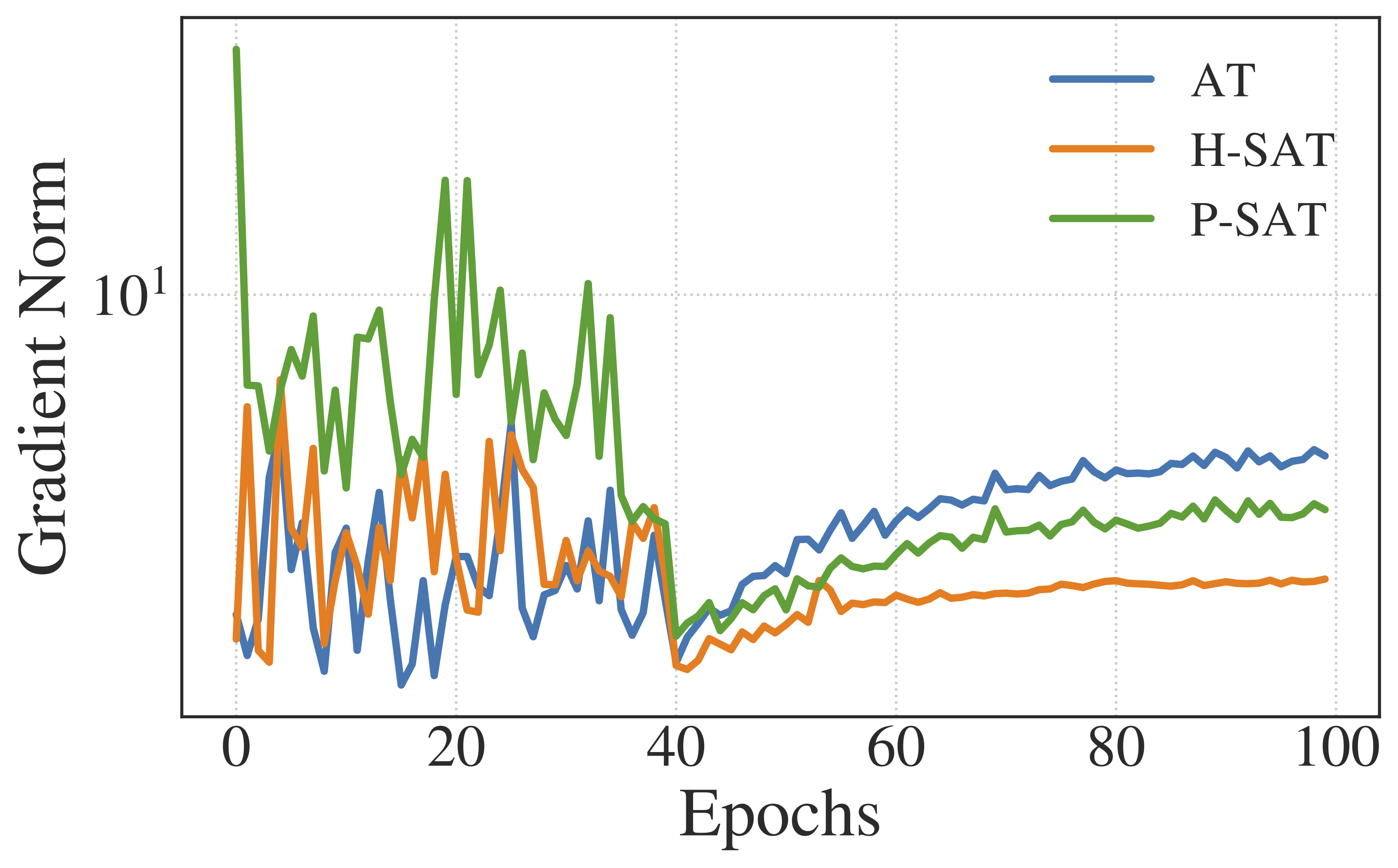}
        \caption{Gradient Norm}
    \end{subfigure}
    \hfill
    \vspace{-5pt}
    \caption{(CIFAR-10) Plots of three quantities typically used to measure the smoothness of the loss landscape of neural networks. Each is calculated exactly at each epoch during the training for the same subset of randomly chosen training samples.}
    \label{fig:smooth}
    \vspace{-10pt}
\end{figure*}

\subsubsection{Larger Improvement with More Difficult Tasks}

On CIFAR-10 with $\epsilon=16/255$ and $24/255$ (Table~\ref{tab:cifar-img}), our schemes have higher clean accuracy compared to \at and higher adversarial accuracy than \trd. 
The difference has increased compared to the experiments with $\epsilon = 8/255$. 
The sum of the two accuracies of \es also continues to beat that of the other defenses.
In the same table, the benefit of our scheme is even more prominent on the more difficult and realistic classification task on Imagenette dataset. 
On this dataset, \hes and \es  outperform the baselines (we omit \cat since it is beaten by the other baselines) by a large margin, even larger than the results on CIFAR-10/100.

The improvement grows larger with an increase in the value of $\epsilon$.
This is because when $\epsilon$ is small, the early termination of PGD is not necessary, and the curriculum does not play a big role.
Hence, \es runs are very similar to those of \at.
Alternatively, this also agrees with the observation made by \citet{bengio_curriculum_2009} and \citet{weinshall_curriculum_2018} that curriculum learning contributes more when the task is more difficult. 
A larger perturbation norm means a stronger adversary and a higher adversarial loss, representing a harder task.

\subsubsection{Smoother Loss Landscape} \label{sssec:smooth}

We empirically confirm that our schemes, both \hes and \es, increase smoothness of the adversarial loss landscape.
To measure smoothness, we plot in Fig.~\ref{fig:smooth} the maximal Hessian eigenvalue, trace of the Hessian, and norm of the gradients on the adversarial examples generated by PGD during training.
There is a significant change in the trend of the three quantities at epoch 40 which is where the learning rate is reduced from 0.05 to 0.005.
SGD almost converges to a local minima at this point, i.e., the loss no longer changes significantly.
Prior to epoch 40, all three quantities behave erratically and vary abruptly.

However, after epoch 40, \hes and \es consistently reach smoother local minima.
Specifically, \es has the smallest eigenvalue and trace of the Hessian. 
\hes is the second smallest while \at comes in last.
For the gradient norm, only \hes and \es switch place, but the general trend remains the same.
This plot is consistent with the visualization in Fig.~\ref{fig:loss} and helps explain why \hes and especially \es perform better than the baselines.

We also measure the generalization gap by computing the difference between training and testing adversarial accuracies over the last 20 epochs of training. 
On CIFAR-10 with $\epsilon=8/255$, the gaps are $2.42 \pm 0.18$, $2.08 \pm 0.21$, and $2.33 \pm 0.24$ for AT, H-SAT, and P-SAT, respectively, where the error denotes a 95\%-confidence interval.
The ordering of the generalization gaps is again consistent with that of spectral norm and trace of the Hessian in Fig.~\ref{fig:smooth}(a), (b).

\section{Conclusion}

We proposed a curriculum-based formulation of adversarial training. 
Keeping the optimization objective unmodified, we focused our analysis on difficulty measures that could guide the model training to smooth regions of the loss landscape, improving generalization. 
Towards this end, we proposed \hes and \es, robust training algorithms that achieved high clean accuracy and a small generalization gap, addressing the two key problems of \at. 
Using extensive evaluation, we showed that \hes and \es outperforms \at, \trd, as well as other curriculum-inspired defenses on both clean and adversarial accuracies in most of the settings.

\begin{acks}
Part of the work done at IBM research was sponsored by the Combat Capabilities Development Command Army Research Laboratory and was accomplished under Cooperative Agreement Number W911NF-13-2-0045 (ARL Cyber Security CRA). The views and conclusions contained in this document are those of the authors and should not be interpreted as representing the official policies, either expressed or implied, of the Combat Capabilities Development Command Army Research Laboratory or the U.S. Government. The U.S. Government is authorized to reproduce and distribute reprints for Government purposes not withstanding any copyright notation here on.

The work done at UC Berkeley was supported by the Hewlett Foundation through the Center for Long-Term Cybersecurity and by generous gifts from Open Philanthropy and Google Cloud Research Credits program with the award GCP19980904.

\end{acks}

\bibliographystyle{ACM-Reference-Format}
\bibliography{reference.bib}

\clearpage

\appendix
We provide below detailed explanations, comparative empirical results, and theoretical analysis of our proposed approach. 
We begin with description of the experiments (datasets, model architecture, training parameters, and algorithm-specific parameters) in Section~\ref{ap:sec:setup}. 
In Section~\ref{ap:sec:constraint}, we detail our curriculum loss framework which unifies the prior works, and then we justify our early termination techniques for \es (Proposition~\ref{prop:early}) and \hes.
Next, in Section~\ref{ap:sec:hessian}, we describe our approximation of the Hessian eigenvalue and additional experiments that compare smoothness of the loss landscape of multiple training schemes.
Lastly, we finish with examples of the adversarial images from the Imagenette dataset in Section~\ref{ap:sec:images}.

\section{Detailed Description of the Experiments}
\label{ap:sec:setup}

The neural networks we experiment with all use ReLU as the activation function and are trained using SGD with a momentum of $0.9$ and batch size of $128$. We use early stopping during training, i.e., models are evaluated at the end of each epoch and only save the one with the highest adversarial validation accuracy thus far. Dataset-specific details of the training as well as brief descriptions of the model architectures are provided below. 

\paragraph{MNIST} All experiments use a three-layer convolution network (8x8-filter with 64 channels, 6x6-filter with 128 channels, and 5x5-filter with 128 channels respectively) with one fully-connected layer. Models are trained for $70$ epochs with a batch size of $128$. The initial learning rate is set at $0.01$ and is decreased by a factor of $10$ at epochs $40, 50,$ and $60$. Weight decay is $5 \times 10^{-4}$. During training, we run PGD for $40$ steps with a step size of $0.02$ and use a uniform random initialization within the $\ell_\infty$-ball of radius $\epsilon$.

\paragraph{CIFAR-10/CIFAR-100} We use both pre-activation ResNet-20~\citep{wang_residual_2017} and WideResNet-34-10~\citep{zagoruyko_wide_2017}, which are trained for $100$ epochs with a batch size of $128$. We use 10-step PGD with step size of $2/255$ and random restart. The initial learning rate is set at $0.05$ for ResNet and $0.1$ for WideResNet and is decreased by a factor of $10$ at epochs $40, 60,$ and $80$. Weight decay is set to $5 \times 10^{-4}$ for CIFAR-10 and $2 \times 10^{-4}$ for CIFAR-100. Standard data augmentation (random crop, flip, scaling, and brightness jitter) is also used.

\paragraph{Imagenette} The hyperparameters are almost identical to those for CIFAR-10. We train ResNet-34 for $100$ epochs with a batch size of $128$. The initial learning rate is $0.1$ and is decreased by a factor of $10$ at epochs $40, 60,$ and $80$. Weight decay is set to $5 \times 10^{-4}$.

For \trd, we set $\beta=6$ for all of the experiments on CIFAR-10, CIFAR-100, and Imagenette, which is the value suggested in the original paper. For MNIST, we did a grid search on the values of $\beta$ including $\beta = 6$. However, we did not find any value of $\beta$ that resulted in a robust model at $\epsilon = 0.45$ on MNIST, so we only report results for $\beta = 6$ in the table.

For \dyn, we followed the schedule used in the original paper, i.e., the convergence score is initialized at $0.5$ and is reduced to zero in $80$ epochs for CIFAR-10 and CIFAR-100. For the other datasets, we found that the adversarial accuracy improved and is more comparable to the other defenses when the convergence score reached zero earlier. So we reduced the number of epochs over which the decay occurred from $80$ to $30$ for MNIST, $50$ for CIFAR-10 with $\epsilon=16/255, 24/255$, and $40$ for Imagenette.

For \cat, we used the original code which is provided by the authors, and we only experimented with the recommended default hyperparameters.

\begin{figure}[t]
    \centering
    \begin{subfigure}[b]{0.235\textwidth}
        \centering
        \includegraphics[width=\textwidth]{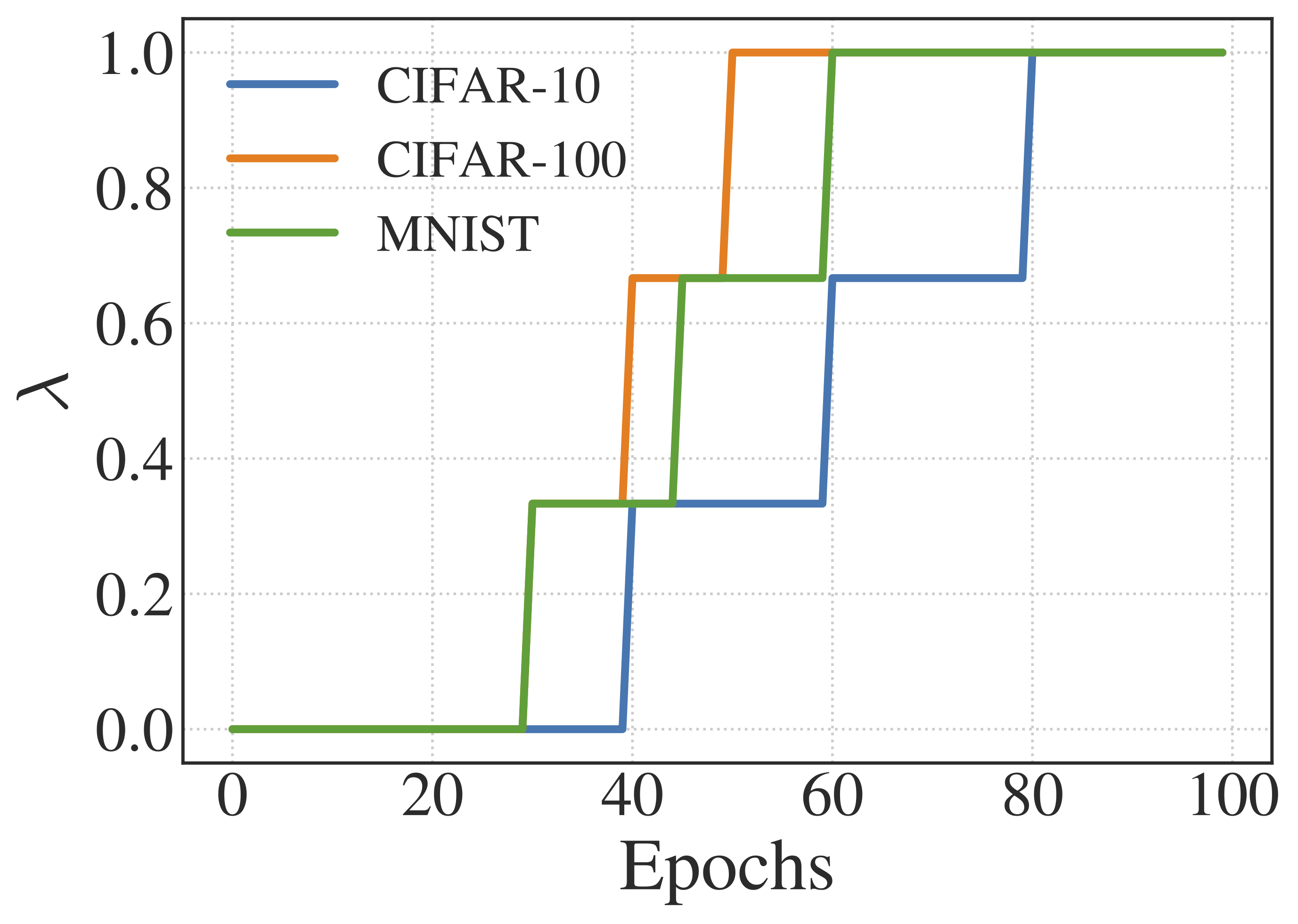}
        \caption{Step Schedules}
    \end{subfigure} \hfill
    \begin{subfigure}[b]{0.235\textwidth}
        \centering
        \includegraphics[width=\textwidth]{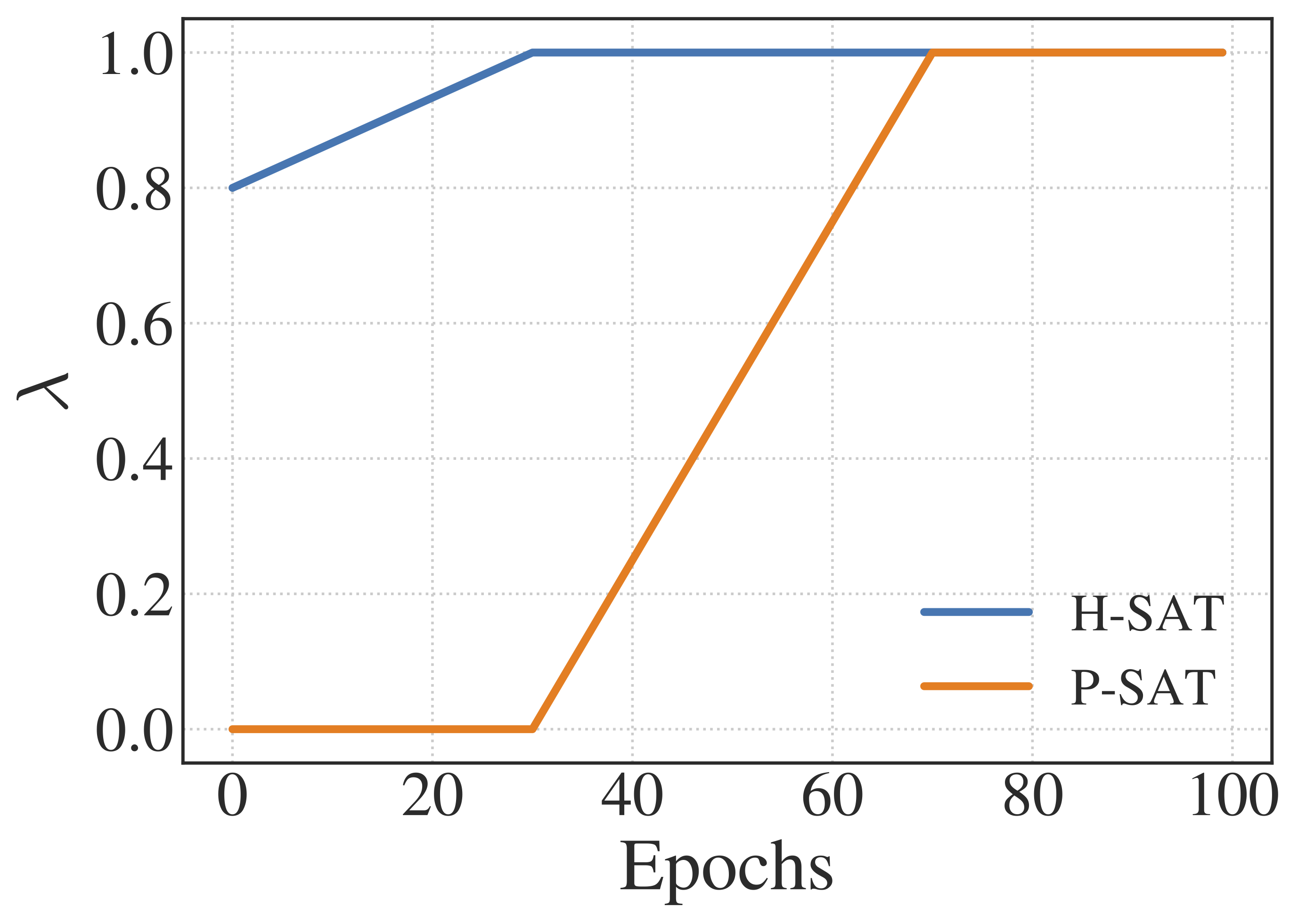}
        \caption{Linear Schedules}
    \end{subfigure}
    \caption{Plots of multiple schedules of the difficulty parameter $\lambda$ used by \hes and \es. (a) displays three step schedules for MNIST, CIFAR-10, and CIFAR-100. (b) shows the linear schedules used in all datasets.}
    \label{fig:schedule}
\end{figure}

For \hes and \es, we used slightly different schedules for the difficulty parameter $\lambda$ depending on the setting (Fig.~\ref{fig:schedule}). For MNIST, we only used a step schedule wherein we increased $\lambda$ is steps of $0.3333$ at epochs $30$, $45$, and $60$. On PRN-20, we used a similar step schedule, i.e., $\lambda$ is increased at epochs $40$, $60$, and $80$. On WRN-34-10, we found that it is beneficial to increase $\lambda$ in earlier epochs. So we increased $\lambda$ at epochs $30$, $40$, and $50$ instead.
In addition to the step schedules, we also use two linear schedules, one for \hes and the other for \es. For \hes, $\lambda$ starts off at 0.8 and increase to 1 by epoch 30. For \es, $\lambda$ increases from 0 to 1 between epoch $30$ and $70$.

All of the codes are written in PyTorch and run on servers with multiple Nvidia 1080ti and 1080 GPUs. On a single Nvidia 1080ti GPU with 12 cores of Intel i7-6850K CPU (3.60GHz) and 64 GB of memory, \es with an MNIST model takes 2 hours to train, PRN-20 takes about 6 hours, and WRN-34-10 uses about 24 hours. \es is slightly faster than \at due to the early termination. On the contrary, \hes uses approximately 50\% additional computation time because of the Hessian eigenvalue computation.

\section{Curriculum Constraints} \label{ap:sec:constraint}

\begin{table*}[t]
\small
\centering
\caption{Comparison of constraints and implementations of different curriculum-based AT schemes unified under the form of Eqn.~\eqref{eq:cl}. Note that \cur uses the number of PGD steps as the constraint which cannot be written analytically.}
\label{tab:metrics}
\begin{tabular}{@{}lll@{}}
\toprule
Schemes & Curriculum Constraints & Implemented Methods \\ \midrule
Perturbation norm & $\norm{\delta}_\infty \le \lambda$ & Projection: $\norm{\delta}_\infty \le \lambda$ \\
\cur~\citep{cai_curriculum_2018} & n/a & Setting PGD steps \\
\dyn~\citep{wang_convergence_2019}    & $\iprod{\delta, \nabla \ell(x+\delta)} - \epsilon \norm{\nabla \ell(x+\delta)}_1 \le \lambda$ & Early termination \\
\iaat~\citep{balaji_instance_2019} & $\psi_P(x+\delta) \le 0$ & Projection: $\norm{\delta}_\infty \le \epsilon^*(x)$ \\
\cat~\citep{cheng_cat_2020} & $\psi_P(x+\delta) \le 0$ & Projection: $\norm{\delta}_\infty \le \epsilon^*(x)$ \\ 
\textbf{\hes (ours)}             & $\psi_H(x+\delta) \le \lambda$ & Subset updates          \\
\textbf{\es (ours)}             & $\psi_P(x+\delta) \le \lambda$ & Early termination          \\ \bottomrule
\end{tabular}
\end{table*}

\subsection{The Curriculum Loss Framework}

Our formulation of the curriculum loss and in particular, the curriculum constraint generalize the curriculum learning approach used by the prior works. Table~\ref{tab:metrics} lists the different curriculum constraints used by each approach together with their method of choice to satisfy the corresponding constraints. \cur controls the difficulty by setting the number of PGD steps for generating adversarial examples. \dyn terminates PGD early once the convergence score is lower than the specified value.

The maximum perturbation norm is also an intuitive difficulty metric that can be scheduled manually or can be automatically adjusted based on some condition. Both \iaat and \cat explicitly use sample-specific perturbation norm, $\epsilon^*(x)$, to control the difficulty. In addition, both methods approximate and schedule $\epsilon^*(x)$ based on the true curriculum constraint (which is designed to keep the perturbed sample on the decision boundary and not push it further inside the boundary of an incorrect class, $\psi_P(x+\delta) \le 0$).

Initially, $\epsilon^*(x)$ is set to a small value. If the perturbed sample is correctly classified, then $\epsilon^*(x)$ increase and adds difficulty for the next epoch. Conversely, if the perturbed sample is incorrectly classified, $\epsilon^*(x)$ is decreased for \iaat or left unchanged for \cat. If this approximation works well, then $\epsilon^*(x)$ should be close to the shortest distance from the decision boundary for each training sample. Also due to the choice of the constraint $\psi_P(x+\delta) \le 0$, \cat to have high accuracy but low robustness. The other difference between \iaat and \cat is that \cat also uses label smoothing which could be the cause of the gradient obfuscation that we observed in our experiments.

\begin{figure}[t]
    \centering
    \includegraphics[width=0.35\textwidth]{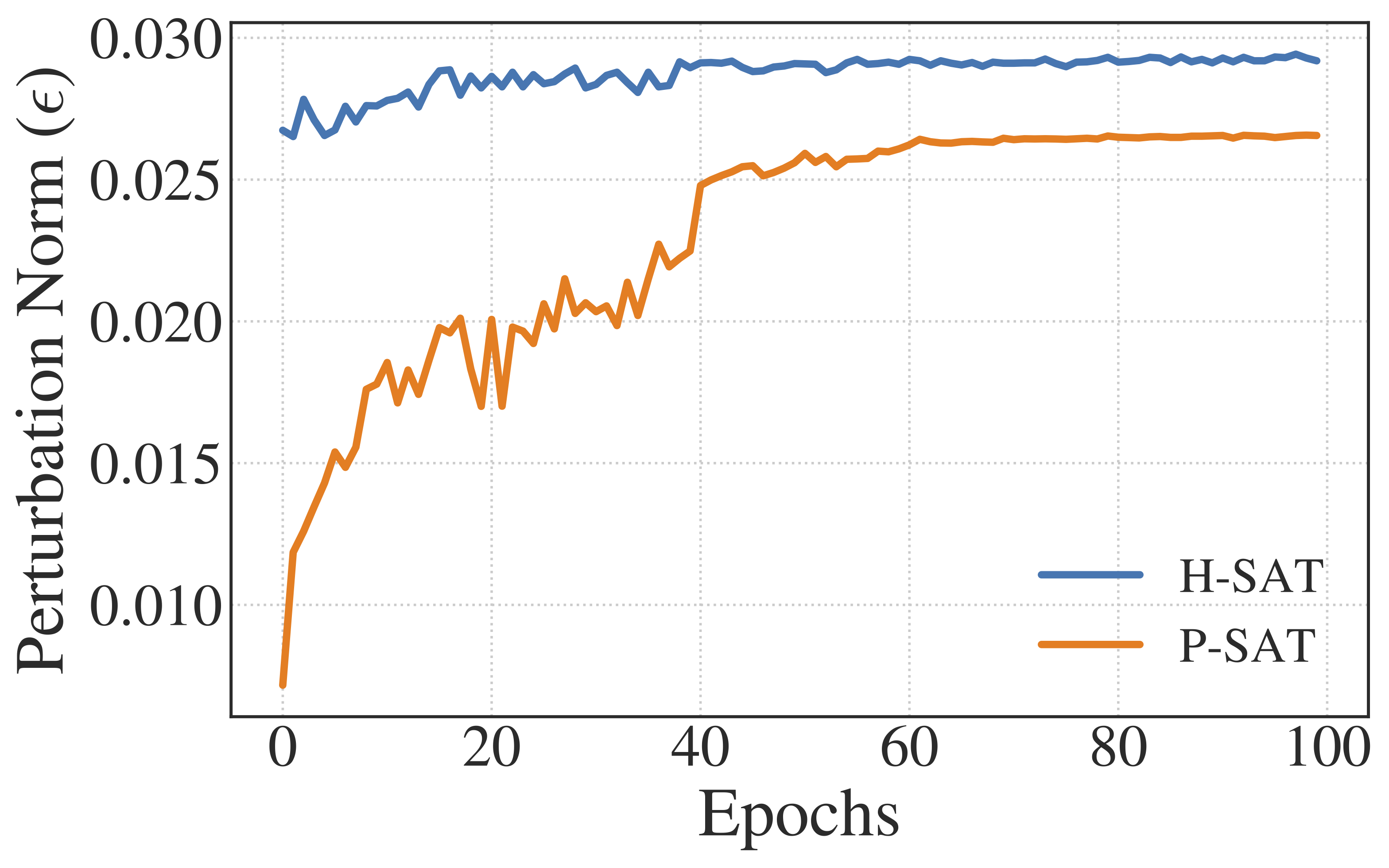}
    \caption{(CIFAR-10) Mean of the perturbation norm ($\epsilon$) of the adversarial examples generated by \hes and \es.  max. $\epsilon$ is set to $8/255$, and no random restart is used to emphasize the trend. Note that \hes and \es use different linear schedules on $\lambda$.}
    \label{fig:pert_norm}
\end{figure}

We claimed that \hes and \es implicitly and adaptively increased the effective perturbation norm as training progressed. Intuitively, the perturbation norm increases for two reasons. First, when the difficulty parameter is increased, the curriculum constraint is relaxed and thus, the samples can be perturbed more on average before the constraint is violated or before PGD is terminated. Second, as the network becomes more robust, a larger perturbation is required to generate adversarial examples with the same level of difficulty. We empirically verify this statement by training networks using \hes and \es on CIFAR-10 with $\epst = 8/255$ and tracking the mean of the perturbation norm for $10$ batches each with $128$ randomly chosen training samples (Fig.~\ref{fig:pert_norm}).

\subsection{Early Termination for \es} \label{ap:ssec:es_early_stop}

Here, we justify the early termination of PGD as a way to solve the curriculum loss for \es.
First, we restate and prove Proposition~\ref{prop:early} for the binary-class case. Then, we extend it to our heuristic for the multi-class case.

\earlystop*

\begin{proof}
First, we rewrite the curriculum objective such that it includes the normal loss function. We consider a fixed $\theta$ here so we drop the dependence on $\theta$ to declutter the notation.
\begin{alignat}{2}
    &\psi_P(x + \delta) \le \lambda \\
    &\iff ~ \max_{j \ne y} f(x + \delta)_j - f(x + \delta)_y &&\le \lambda \\
    &\iff ~ 1 - f(x + \delta)_y - f(x + \delta)_y &&\le \lambda \\
    &\iff ~ f(x + \delta)_y &&\ge (1 - \lambda)/2 \\
    &\iff ~ - \log\left( f(x + \delta)_y \right) &&\le - \log\left( (1 - \lambda)/2 \right) \\
    &\iff ~ \ell(x + \delta) &&\le \gamma \\
    &\qquad\quad \gamma \coloneqq - \log\left( (1 - \lambda)/2 \right)
\end{alignat}
Now we can rewrite the optimization in Eqn.~\eqref{eq:cl} as
\begin{align}
    \max_{\delta:\norm{\delta}_p \le \epst} ~&\ell(x+\delta) & &= &\max_{\delta:\norm{\delta}_p \le \epst} ~&\ell(x+\delta) \\
    \text{s.t.} \quad &\psi_P(x+\delta) \le \lambda & & & \text{s.t.} \quad &\ell(x + \delta) \le \gamma \\
    \quad & \quad & &= & \max_{\delta:\norm{\delta}_p \le \epst} &\min\left\{ \ell(x+\delta), \gamma \right\}
\end{align}
Now we have arrived at the modified optimization problem that only has one convex constraint on the norm of $\delta$. This problem can then be solved with PGD (we call it PGD, but it is a projected gradient ascent). Given a particular sample $(x, y)$, let's consider two possible cases. First, $\nexists \delta$ such that $\norm{\delta}_\infty \le \epsilon$ and $\ell(x+\delta) \ge \gamma$. In this case, the new problem reduces to the adversarial loss whose local optima can be found by PGD with a sufficient number of iterations. 

The second scenario, $\exists \delta$ such that $\norm{\delta}_\infty \le \epsilon$ and $\ell(x+\delta) \ge \gamma$, can be solved if we can find any $\delta$ that satisfy the two conditions. Since projected gradient methods satisfy the first condition by default, we only have to run PGD until the second condition is satisfied. Equivalently, PGD can be terminated as soon as the curriculum constraint is violated.
\end{proof}

For the multi-class case, $\gamma$ is dependent on $\delta$ so Eqn.~\eqref{eq:cl} does not reduce to a similar simple form.
\begin{align}
    &\psi_P(x + \delta) \le \lambda \\
    &\iff ~ \max_{j \ne y} f(x + \delta)_j - f(x + \delta)_y \le \lambda \\
    &\iff ~ f(x + \delta)_y \ge \max_{j \ne y} f(x + \delta)_j - \lambda \\
    &\iff ~ - \log\left( f(x + \delta)_y \right) \le - \log\left( \max_{j \ne y} f(x + \delta)_j - \lambda \right) \\
    &\iff ~ \ell(x + \delta) \le \gamma(x+\delta, \lambda) \\
    &\qquad \gamma(x+\delta, \lambda) \coloneqq - \log\left( \max_{j \ne y} f(x + \delta)_j - \lambda \right)
\end{align}
Note that we can assume that the RHS on line 2, $\max_{j \ne y} f(x + \delta)_j - \lambda$, is positive. Otherwise, the constraint is automatically satisfied and can be ignored because the LHS $f(x + \delta)_y$ is always non-negative. Similarly to the binary-class case, Eqn.~\eqref{eq:cl} can be rewritten as:
\begin{align}
    \max_{\delta:\norm{\delta}_p \le \epst} ~&\ell(x+\delta) \\
    \text{s.t.} \quad &(x+\delta) \le \lambda \\
    = \max_{\delta:\norm{\delta}_p \le \epst} ~&\ell(x+\delta) \\ \text{s.t.} \quad &\ell(x + \delta) \le \gamma(x+\delta, \lambda) \\
    = \max_{\delta:\norm{\delta}_p \le \epst} &\min\left\{ \ell(x+\delta), \gamma(x+\delta, \lambda) \right\}
\end{align}

This objective is difficult to optimized in a few steps of PGD because of the piecewise min and max as well as the fact that the two terms are inversely proportional to the other.
Alternatively, we propose a heuristic to approximate the objective by treating the second term as a constant and so not computing its gradients (we still update it as $\delta$ changes).

\subsection{Early Termination for \hes} \label{ap:ssec:hes_early_stop}

The early termination can also be applied in this case. However, the approximation of the max eigenvalue of the Hessian vary significantly across models and datasets. Thus, setting a hard threshold is cumbersome and requires a lot fine-tuning.
Instead, we propose the following scheme that leverages the high rank-correlation between the true and the approximated eigenvalue to circumvent the above issue of setting hard thresholds.

We choose $\lambda$ as fraction of samples with the smallest Hessian eigenvalue from each batch to perturb in a given PGD step. For example, when $\lambda = 0.3$, only 30\% of the samples in the batch with the smallest $\norm{H_\epsilon(x,\theta)}_2$ are perturbed. This method adapts to samples in the batch and is more flexible than a hard threshold.
Now, similarly to the probability gap, $\lambda$ starts off with a small value and increases to $1$, which is equivalent to \at, at the end of training.
Setting $\lambda = 0$ is equivalent to normal training as no samples are perturbed.
\section{Maximal Hessian Eigenvalue} \label{ap:sec:hessian}

\subsection{Maximal Eigenvalue Approximation} \label{ap:ssec:hess_approx} 

\begin{figure*}[ht!]
    \centering
    \hfill
    \begin{subfigure}[b]{0.25\textwidth}
        \centering
        \includegraphics[width=\textwidth]{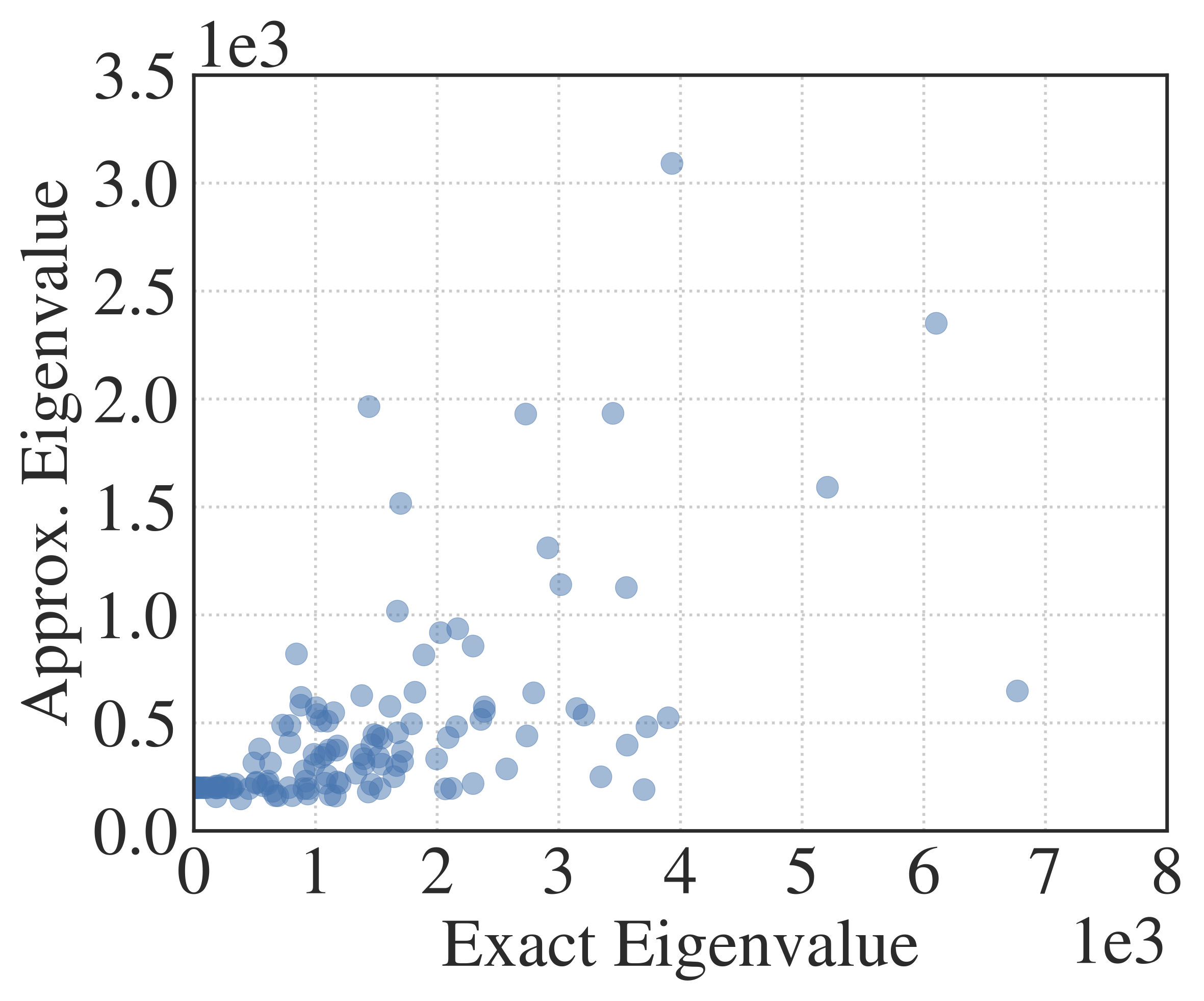}
        \caption{\at}
    \end{subfigure} 
    \hfill
    \begin{subfigure}[b]{0.25\textwidth}
        \centering
        \includegraphics[width=\textwidth]{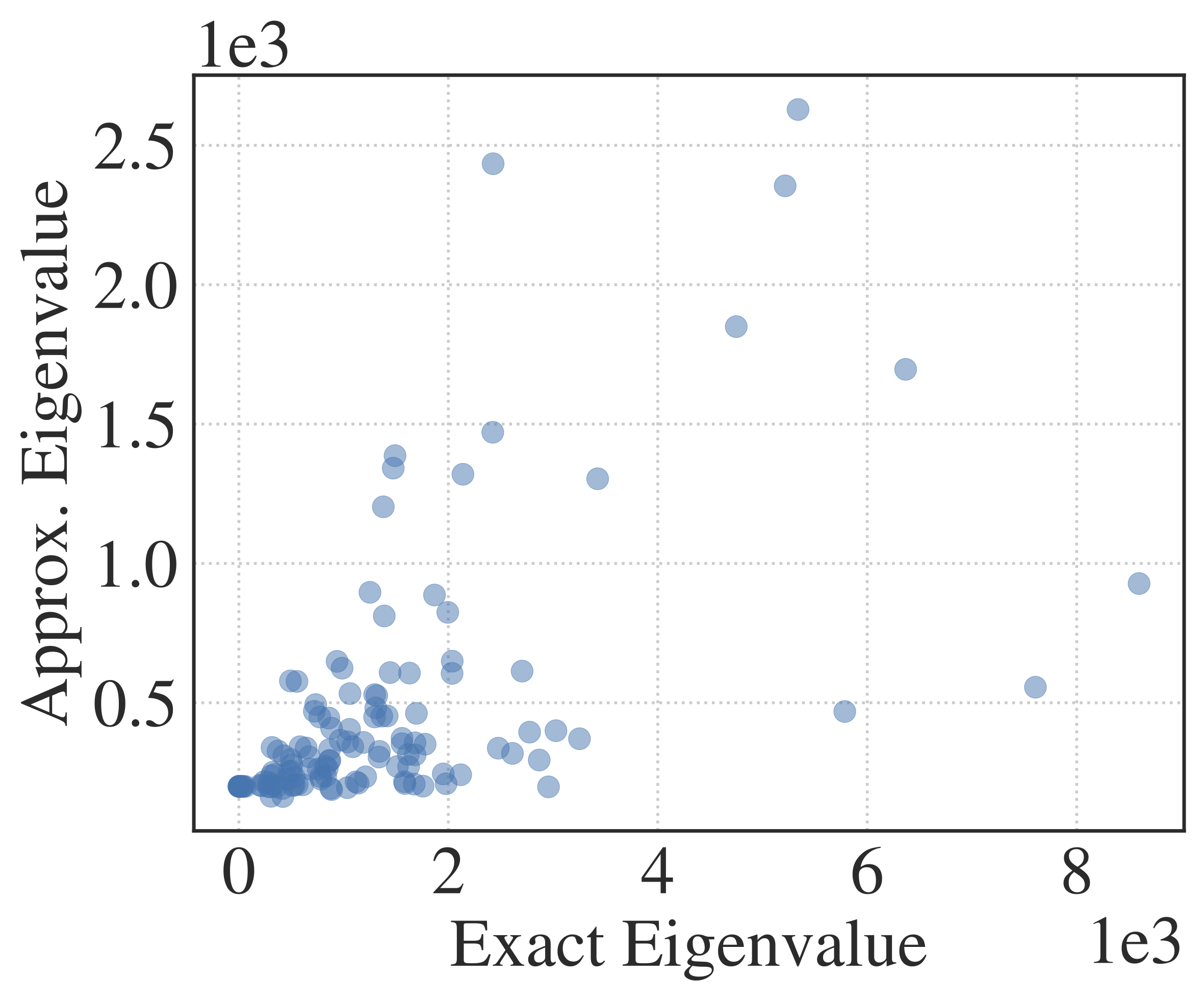}
        \caption{\hes}
    \end{subfigure}
    \hfill
    \begin{subfigure}[b]{0.25\textwidth}
        \centering
        \includegraphics[width=\textwidth]{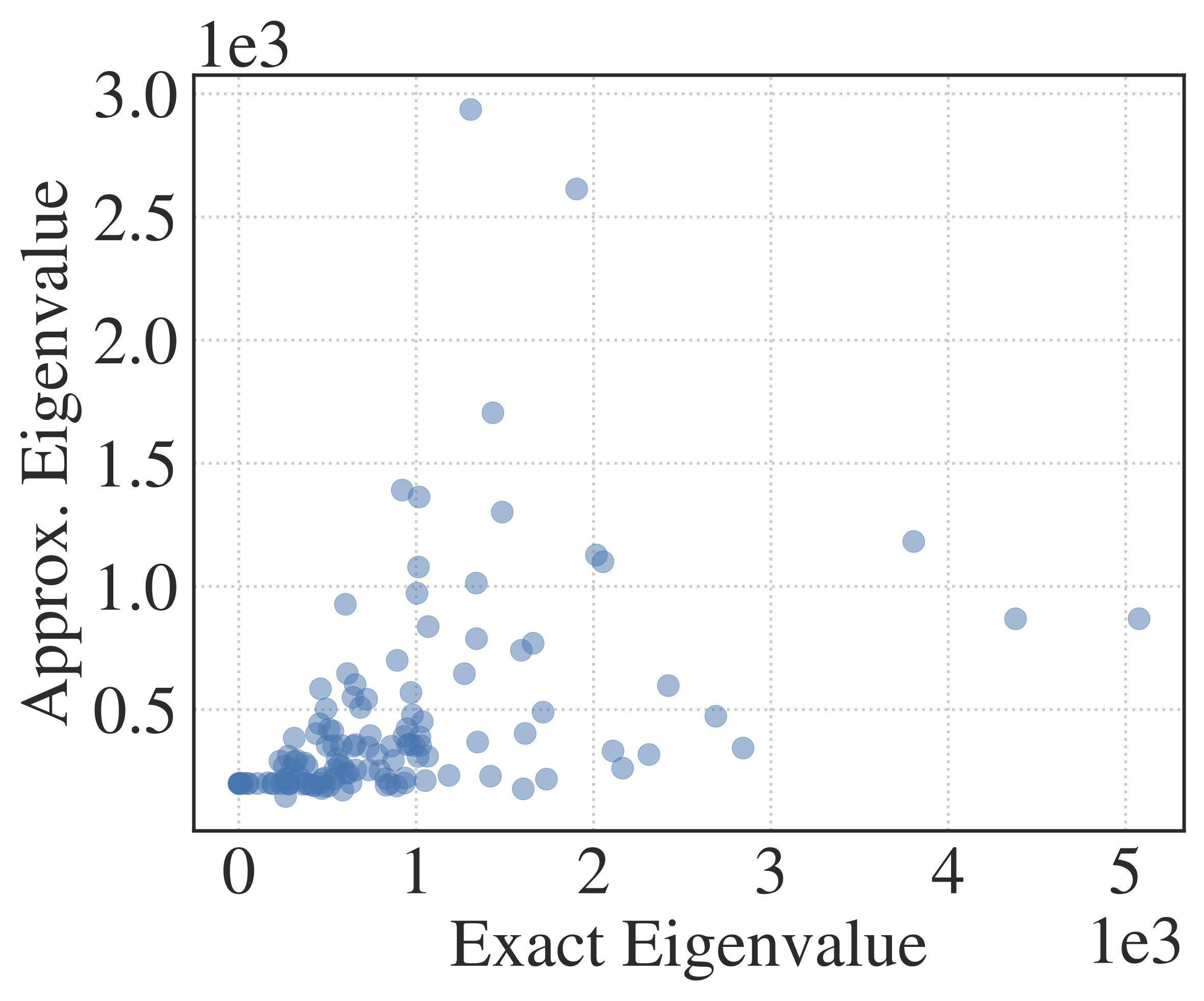}
        \caption{\es}
    \end{subfigure}
    \hfill\phantom{~}
    \caption{(CIFAR-10) The exact and the approximate maximal eigenvalue of the Hessian computed on three ResNet models: (a) \at, (b) \hes, and (c) \es. Each point corresponds to one randomly chosen training sample.}
    \label{fig:eigen_correlation}
\end{figure*}

Below we outline a series of approximations for minimizing the overhead of computing the max. Hessian eigenvalue.
We use the second-order Taylor's expansion to approximate the Hessian:
\begin{align}
    \scalemath{0.92}{
    \ell(\theta + \zeta) = \ell(\theta) + \zeta^\top \nabla_\theta \ell(\theta) + \frac{1}{2}\zeta^\top \nabla^2_\theta \ell(\theta) \zeta + \mathcal{O}(\norm{\zeta}_2^3)
    }
\end{align}
Note, we can now maximize over $\zeta$ to obtain the max. eigenvalue of the Hessian. 
The absolute value can be omitted because we are only concerned with the positive eigenvalues.
\begin{align}
    \frac{1}{2} &\norm{\nabla^2_\theta \ell(\theta)}_2 = \max_{\norm{\zeta}_2 = 1} \frac{1}{2}\zeta^\top \nabla^2_\theta \ell(\theta) \zeta \\
    &= \max_{\norm{\zeta}_2 = 1} \left[\ell(\theta + \zeta) - \zeta^\top \nabla_\theta \ell(\theta) \right] - \ell(\theta) \label{eq:approx_hess} \\
    &\le \max_{\norm{\zeta}_2 = 1} \ell(\theta + \zeta) - \min_{\norm{\zeta}_2 = 1} \zeta^\top \nabla_\theta \ell(\theta) - \ell(\theta) \label{eq:approx_hess_ineq}\\
    &\approx \frac{1}{\alpha} \left\{ \ell(\theta + \alpha g) + \alpha \norm{\nabla_\theta \ell(\theta)}_2 \right\} - \ell(\theta)  \label{eq:ub}
\end{align}
where $g$ is a shorthand notation of $g(x;\theta) \coloneqq \nabla_\theta \ell(x;\theta) / \norm{\nabla_\theta \ell(x;\theta)}_2$. 
The first and the second terms in Eqn.~\eqref{eq:ub} approximate the maximization and the minimization in Eqn.~\eqref{eq:approx_hess_ineq} by taking a one-step projected gradient update.
We can also use a similar approximation to lower bound Eqn.~\eqref{eq:approx_hess} by substituting $\zeta$ with $\alpha g$ and $-\alpha g$ and take the maximum between the two:
\begin{align}
    \frac{1}{2} \norm{\nabla^2_\theta \ell(\theta)}_2 ~\gtrapprox&~ \frac{1}{\alpha}\max \{ \ell(\theta + \alpha g) - \alpha \norm{\nabla_\theta \ell(\theta)}_2, \label{eq:lb} \\ 
    &\ell(\theta - \alpha g) + \alpha \norm{\nabla_\theta \ell(\theta)}_2\} - \ell(\theta) \nonumber
\end{align}
Now we have arrived at the upper and the lower bounds as stated in Section~\ref{sec:compute}. 
Evaluating Eqns.~\eqref{eq:ub} and \eqref{eq:lb} at the adversarial example of $x$, gives us the upper and a lower bound respectively, for $\norm{H_\epsilon(x,\theta)}_2$.
Choosing an appropriate value of $\alpha$ makes the Taylor's series based approximation and the maximization using a single gradient step much more accurate in practice. 
We choose it to be 1\% of the gradient norm so the precision also automatically adapts to the current scale.
Choosing $\alpha$ much smaller is not recommended because it can blow up small numerical errors

\subsection{Implementation Consideration} \label{ap:ssec:practice}

Note that the smoothness analysis typically computes eigenvalue of the Hessian matrix of the loss averaged over the entire training set, but for the purpose of curriculum learning, we want to control the Hessian eigenvalue for individual samples.
The former quantity can be upper bounded by the average of the latter as follows:
\begin{align}
    \norm{\frac{1}{n} \sum_{i=1}^n \nabla^2_\theta \ell_\epsilon(x_i;\theta)}_{(2)} \le \frac{1}{n} \sum_{i=1}^n \norm{H_\epsilon(x_i, \theta)}_{(2)}
\end{align}
This shows that Hessian eigenvalue can be used as a difficulty metric for curriculum learning and still controls the smoothness of the loss landscape.

We have derived the lower/upper bounds of the maximal Hessian eigenvalue in Section~\ref{sec:compute} and Appendix~\ref{ap:ssec:hess_approx}. 
Nonetheless, we face with some difficulty for combining it with \at in practice.
To enable a fine-grained sample-wise control on the difficulty metric, we must approximate the Hessian eigenvalue per sample at every PGD step of \at.
This is an issue in practice as the automatic differentiation software (e.g., PyTorch) does not provide an easy way to access $\nabla_\theta \ell(x; \theta)$, and evaluation of $\ell(x; \theta + \nabla_\theta \ell(x; \theta))$ cannot be parallelized due to the fact that gradients are different for each sample $x$. 
If we ignore the parallelization and compute the perturbed loss for every $x$ in the batch in a sequential manner, the computation time becomes prohibitively expensive (linear in minibatch size).

To reduce the computation, we approximate gradients of individual samples with the minibatch gradient.
Obviously, if the minibatch size were set to one, this approximation is exact.
The smaller the minibatch size, the more accurate and more expensive this gradient approximation becomes. 
However, we want to keep the minibatch size fixed across all the defenses we experiment with for a fair comparison.
Thus, we avoid the issue by fixing the minibatch size for the weight update to be 128 (same as the other schemes) but using a smaller minibatch size for computing the Hessian eigenvalue.

We determine that a minibatch size of 32 for the Hessian computation is sufficiently accurate and does not introduce too much overhead.
We measure the Spearman rank-correlation for the samplewise Hessian eigenvalue computed exactly by the power method and approximately by the upper bound and the heuristic we introduced above.
The correlation is above 0.6 in all the cases we test.
In Fig.~\ref{fig:eigen_correlation}, we plot the eigenvalue for 10 batches each with 128 randomly chosen training samples computed on three ResNet's trained with \at, \hes, and \es. The correlations are 0.6419, 0.6097, and 0.6456 for the three models respectively.

\section{Examples of Adversarial Images} 
\label{ap:sec:images}
\begin{figure*}[ht!]
\centering
\includegraphics[width=\textwidth]{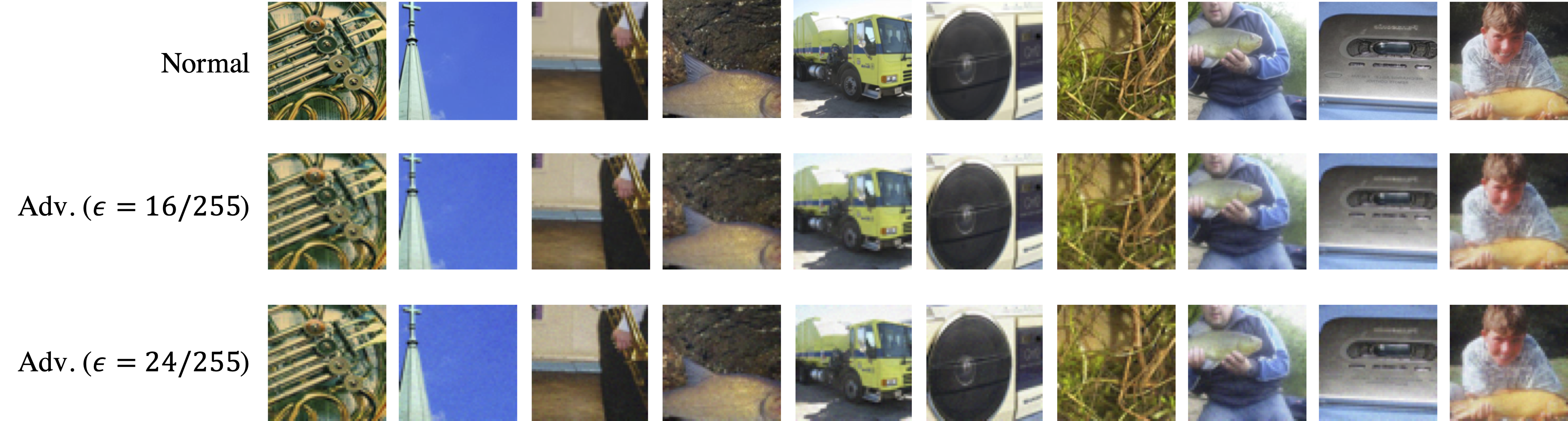}
\caption{(Imagenette) Randomly selected images from the Imagenette dataset. The first row is the original images while the second and the third are adversarial examples generated with $\epsilon$ of $16/255$ and $24/255$ respectively.}
\label{fig:images}
\end{figure*}

Fig.~\ref{fig:images} shows 10 examples of the images from the Imagenette dataset, which is a ten-class subset of the ImageNet dataset.
The images in the second and the third rows are adversarial examples that are perturbed with $\epsilon$ of $16/255$ and $24/255$ respectively.
This figure illustrates that while the choices of perturbation norm we experiment with may seem large compared to $8/255$ for CIFAR-10/100, they are very much imperceptible to humans because the images are of much higher resolution (224 by 224 pixels).

\end{document}